\documentclass{article}

\usepackage{microtype}
\usepackage{graphicx}
\usepackage{subcaption}
\usepackage{booktabs} % for professional tables

\usepackage{hyperref}

\usepackage[preprint]{icml2026}

\usepackage{amsmath}
\usepackage{amssymb}
\usepackage{mathtools}
\usepackage{amsthm}

\usepackage[capitalize,noabbrev]{cleveref}

\theoremstyle{plain}

\theoremstyle{definition}

\theoremstyle{remark}

\usepackage[textsize=tiny]{todonotes}

\icmltitlerunning{AGPO: Asymmetric Group Policy Optimization for Verifiable Reasoning and Search Ads Relevance at JD}

\usepackage{tabularx}
\usepackage{booktabs}
\usepackage[table]{xcolor} 
\newcolumntype{Y}{>{\centering\arraybackslash}X} 
\definecolor{rowgray}{gray}{0.96}
\definecolor{rowblue}{HTML}{c7e7f2} 
\usepackage{CJKutf8}

\usepackage{enumitem} 
\usepackage{natbib}   
\usepackage{tcolorbox}
\usepackage[utf8]{inputenc}

\usepackage{multirow}

\begin{document}

\twocolumn[
  \icmltitle{AGPO: Asymmetric Group Policy Optimization for Verifiable Reasoning \\ and Search Ads Relevance at JD}

  % It is OKAY to include author information, even for blind submissions: the
  % style file will automatically remove it for you unless you've provided
  % the [accepted] option to the icml2026 package.

  % List of affiliations: The first argument should be a (short) identifier you
  % will use later to specify author affiliations Academic affiliations
  % should list Department, University, City, Region, Country Industry
  % affiliations should list Company, City, Region, Country

  % You can specify symbols, otherwise they are numbered in order. Ideally, you
  % should not use this facility. Affiliations will be numbered in order of
  % appearance and this is the preferred way.
  \icmlsetsymbol{equal}{*}

  \begin{icmlauthorlist}
    \icmlauthor{Yang Xu}{nju}
    \icmlauthor{Kun Yao}{pku}
    \icmlauthor{Yiming Deng}{jd}
    \icmlauthor{Zheng Fang}{jd}
    \icmlauthor{Kai Ming Ting}{nju}
    \icmlauthor{Ming Pang}{jd}
  \end{icmlauthorlist}

  \icmlaffiliation{nju}{Nanjing University, Nanjing, China}
  \icmlaffiliation{pku}{Peking University, Beijing, China}
  \icmlaffiliation{jd}{JD.com, Beijing, China}

  \icmlcorrespondingauthor{Yang Xu}{xuyang@lamda.nju.edu.cn}
  \icmlcorrespondingauthor{Ming Pang}{pangming8@jd.com}

  % You may provide any keywords that you find helpful for describing your
  % paper; these are used to populate the "keywords" metadata in the PDF but
  % will not be shown in the document
  \icmlkeywords{Reinforcement Learning, LLM, Reasoning, RLVR}

  \vskip 0.3in
]

% this must go after the closing bracket ] following \twocolumn[ ...

\printAffiliationsAndNotice{} % or {}

% \begin{abstract}
%     Reinforcement Learning with Verifiable Rewards (RLVR) has demonstrated notable success in enhancing the reasoning performance of Large Language Models (LLMs). However, recent studies reveal that while current RLVR methods (e.g., PPO, GRPO) improve sampling efficiency towards correct paths, they do not elicit fundamentally new reasoning patterns. Instead, the reasoning capability boundary of trained models often narrows compared to their base models, with base models achieving higher coverage at large sample sizes ($k$). In this work, we propose Asymmetric Group Policy Optimization (AGPO) to counteract this boundary shrinkage. AGPO adopts a dominant negative reinforcement strategy to suppress incorrect reasoning paths, thereby maintaining the base model's exploration capacity. Crucially, for positive reinforcement, AGPO introduces a noise-gated group advantage mechanism. This mechanism dynamically modulates the update magnitude based on intra-group variance, allowing the model to selectively reinforce samples with high information gain while mitigating instability from low-variance successes. Experiments on MATH benchmarks demonstrate that AGPO achieves state-of-the-art accuracy while consistently improving pass@k performance at scale. Furthermore, in a large-scale industrial application for Search Ads Relevance Optimization, AGPO significantly enhances the quality of synthesized training data, leading to substantial performance gains in downstream student models.
% \end{abstract}

\begin{abstract}
    Reinforcement Learning with Verifiable Rewards (RLVR) has demonstrated notable success in enhancing the reasoning performance of large language models (LLMs). However, recent studies reveal that while current RLVR methods improve sampling efficiency towards correct paths, they do not elicit fundamentally new reasoning patterns. Instead, the reasoning capability boundary of trained models often narrows compared to their base models, with base models achieving higher coverage at large sample sizes. In this work, we propose \textbf{A}symmetric \textbf{G}roup \textbf{P}olicy \textbf{O}ptimization (AGPO) to counteract this boundary shrinkage. AGPO adopts a negative-dominant reinforcement strategy to suppress incorrect reasoning paths, maintaining the base model's exploration capacity. For positive reinforcement, AGPO adopts a group advantage mechanism, which scales positive updates based on intra-group variance, allowing the model to focus on rare correct paths while suppressing updates from trivial paths. Our experiments on five \textit{mathematical benchmarks} demonstrate that AGPO achieves state-of-the-art accuracy while consistently improving pass@$k$ performance at scale. In a large-scale industrial application for \textit{search ads relevance optimization}, AGPO effectively enhances the quality of the data annotation, leading to substantial performance gains in downstream student models.
\end{abstract}

\section{Introduction}

Post-training with reinforcement learning has established itself as the standard paradigm for developing reasoning-centric large language models (LLMs), adopted by frontier models such as OpenAI-o1~\cite{jaech2024openai}, DeepSeek-R1~\cite{guo2025deepseek}, and Qwen3~\cite{yang2025qwen3}. This paradigm has significantly advanced LLMs' capabilities in tackling complex logical tasks involving mathematics and programming, where Reinforcement Learning with Verifiable Rewards (RLVR) serves as a critical engine. Starting from a pretrained base model or one fine-tuned on long Chain-of-Thought (CoT) data, RLVR optimizes the policy via Reinforcement Learning (RL) guided by simple, automatically computable rewards. Through this simple but effective framework, RLVR overcomes the ambiguity and sparsity inherent in subjective human feedback~\cite{shao2024deepseekmath, lambert2024tulu}. Mirroring the success of traditional RL in domains such as gaming and autonomous driving~\cite{mnih2015human, silver2017mastering,sallab2017deep}, where agents autonomously discover new strategies to surpass human performance, it is widely believed that RLVR similarly empowers LLMs to evolve emergent reasoning patterns, thus transcending the inherent capabilities of their base models~\cite{guo2025deepseek}.

However, empirical reality tells a different story. 
Recent studies reveal that while standard RL algorithms adopted for RLVR (\textit{e.g.}, PPO~\cite{schulman2017proximal}, GRPO~\cite{shao2024deepseekmath} and REINFORCE~\cite{williams1992simple}) significantly improve sampling efficiency, they typically fail to elicit fundamentally new reasoning paths~\cite{yue2025does, zhu2025surprising, liu2025reinforcement}.
These algorithms effectively bias the model towards correct paths found in the base distribution, but this often comes at the cost of the reasoning capability boundary.
Specifically, although single-turn accuracy (Pass@$1$) improves, the model's coverage of solvable problems at large sample sizes (Pass@$k$) even degrades compared to the base model~\cite{yue2025does}. 
This implies that the mechanism of RLVR is to improve the probability of correct paths already latent in the base model, rather than generating new correct paths.

With this in mind, one feasible strategy is distillation from stronger teacher models, which is to introduce new reasoning paths to improve the model's capabilities~\cite{yue2025does, hu2025distillation}.
However, it inevitably incurs substantial computational overhead and dependence on pre-trained teacher models with significantly larger parameters~\cite{guo2025deepseek}. Moreover, for frontier models scaling to hundreds of billions of parameters, a superior teacher is often unavailable, rendering self-evolution via RL still a critical and prevalent pathway~\cite{jaech2024openai,yang2025qwen3}.
In fact, current RLVR methods have not sufficiently utilized the latent potential of base models.
Even if the base model can cover the vast majority of correct paths at sufficiently large sample sizes (Pass@$k$), the accuracy of optimized models remains far below this latent upper bound, often at the cost of a shrunken reasoning boundary~\cite{yue2025does,liu2025reinforcement}. This highlights the current challenge of RLVR: how to improve the model's ability towards the base model's optimal coverage while preserving the reasoning boundary?
Fortunately, a recent work based on W-REINFORCE~\cite{zhu2025surprising} shows that improving model accuracy without sacrificing the reasoning boundary is achievable through rebalancing learning signals from Positive Sample Reinforcement (PSR) and Negative Sample Reinforcement (NSR). This work shows that the NSR component serves as a safer optimization driver, allowing the model to improve capabilities by pruning incorrect paths, rather than forcing the distribution to collapse onto a few correct paths.

Motivated by these recent insights on RLVR, we propose Asymmetric Group Policy Optimization (AGPO) to address the limitations of current RLVR algorithms. AGPO is driven by the following three core strategies.
\begin{enumerate}[label=(\arabic*)]
    \item \textit{Dominant negative reinforcement}. Unlike PSR which tends to sharpen the output distribution, NSR functions as a \textit{soft reranking} mechanism that suppresses incorrect paths to redistribute probability mass toward valid priors, thereby improving the base model without collapsing the reasoning capability boundary~\cite{zhu2025surprising}. However, exclusive reliance on such implicit redistribution can compromise training stability and greedy decoding accuracy. This motivates a strategy for RLVR in which NSR serves as the primary driver, supplemented by PSR only when necessary.
    \item \textit{Zero intervention for the trivial correct paths}. Empirical evidence reveals that the correct paths in RLVR training are typically already latent in the base model, and continuously reinforcing them leads to a shrunken reasoning boundary~\cite{yue2025does}. This implies that updates for such trivial correct paths should be excluded from PSR to prevent the distribution from excessively sharpening towards simple paths at the expense of rare, complex but correct reasoning paths.
    \item \textit{Higher rewards for rarer correct paths}. The high variance of policy gradient methods requires the estimation of advantages using a baseline, such as the group average, to measure relative advantages among the sampled trajectories~\cite{shao2024deepseekmath}. This strategy is critical for PSR in RLVR, as it enables the effective improvement of the probability of latent rare correct paths within the base model, thereby enhancing the trained model's greedy reasoning capability.
\end{enumerate}

\begin{figure}[t]
    \centering
    \includegraphics[width=\linewidth]{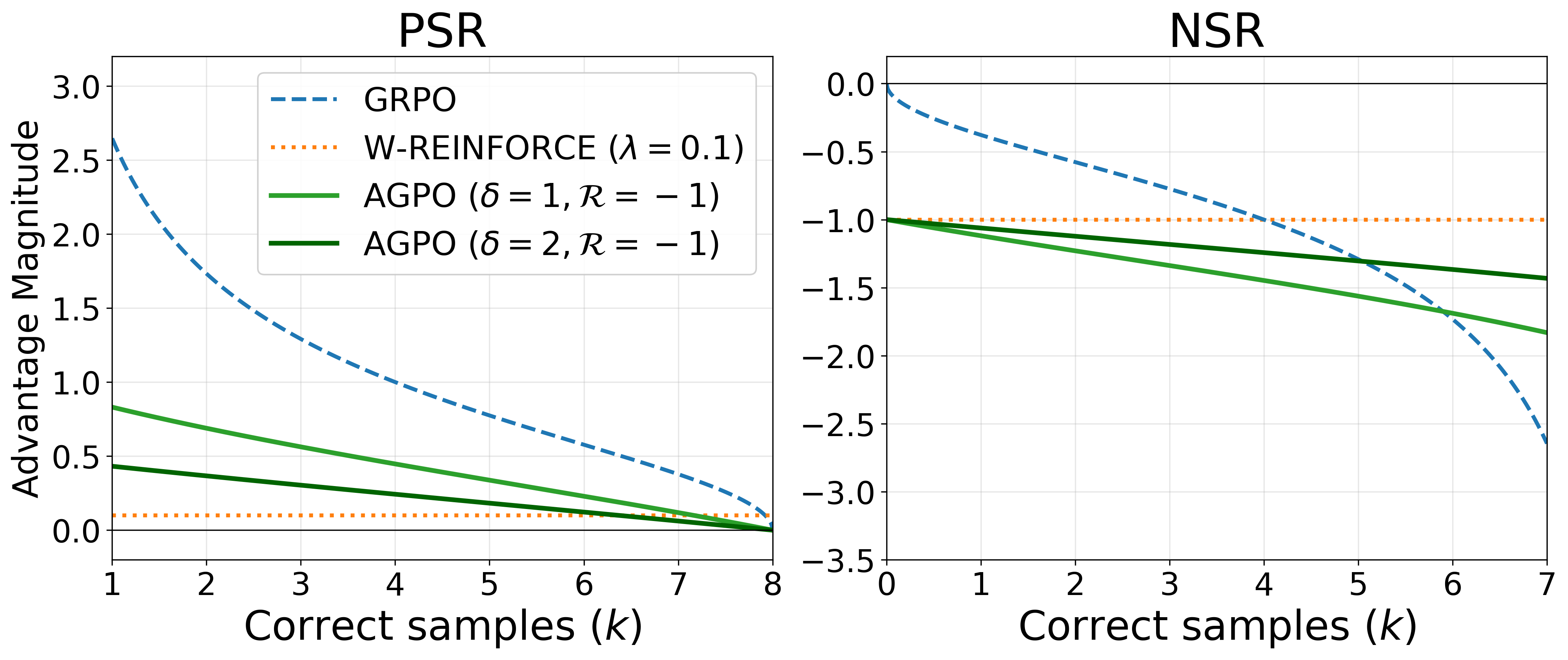}
    \caption{Advantage magnitude comparison: AGPO vs. baselines. AGPO employs asymmetric dynamic advantage estimation, with advantages in both PSR and NSR showing nearly linear trends.}
    \label{fig:reward_dynamics}
\end{figure}

Figure~\ref{fig:reward_dynamics} illustrates the advantage magnitudes (rewards) of AGPO compared to GRPO and W-REINFORCE. 
First, both AGPO and W-REINFORCE adopt an NSR-dominant strategy, where the magnitude of negative penalties significantly outweighs that of positive rewards. Second, both AGPO and GRPO dynamically adjust the advantages based on the number of correct samples ($k$) within the group. Notably, AGPO exhibits a nearly linear response to $k$, providing stable gradient signals that decrease as the consensus of the group increases.

Our main contributions are as follows:
\begin{itemize}
    \item Proposing AGPO that uses the NSR-dominated dynamic advantage estimation to overcome the limitations of current RLVR algorithms.
    
    \item Showing the effectiveness of AGPO measured by a range of Pass@$k$ metrics through five mathematical reasoning benchmarks including \texttt{MATH}, \texttt{Olympiad}, \texttt{AIME-2025}, \texttt{AIME-2024}, and \texttt{GaoKao-2024}.
    
    \item Validating the practical value of AGPO through a real-world industrial task, \textit{search ads relevance optimization}, to enhance the quality of the data annotation.
\end{itemize}

\section{Preliminaries}
\label{sec:pre}

In this section, we begin by reviewing the fundamentals of RLVR and key evaluation metrics, then synthesize empirical insights from recent work.

\subsection{Reinforcement Learning with Verifiable Rewards}

Reinforcement Learning with Verifiable Rewards (RLVR) has emerged as a dominant paradigm for enhancing the reasoning capabilities of LLMs, particularly in domains such as mathematics and code generation, where outcomes are objectively measurable~\cite{jaech2024openai,shao2024deepseekmath}. 
Central to this framework is a deterministic verification function $\mathcal{V}(x,\mathbf{y})$, which evaluates the correctness of a model-generated response (\textit{i.e.}, a token sequence) $\mathbf{y} = (y_1, \dots, y_T)$ given a natural-language prompt $x$ sampled from the distribution of prompts $\mathcal{D}$. 
Unlike Reinforcement Learning from Human Feedback (RLHF)~\cite{ouyang2022training,gao2023scaling}, which relies on subjective and costly human annotations, the verifier $\mathcal{V}(x,\mathbf{y})$ assigns a sparse, binary reward $r = \mathcal{V}(x,\mathbf{y})\in\{0,1\}$ based solely on the validity of the final answer.
Formally, the fundamental objective of RLVR is to optimize the policy parameters $\theta$ to maximize the expected reward over the data distribution:
\begin{equation}
    \mathcal{J}(\theta) = \mathbb{E}_{x \sim \mathcal{D}} \left[ \mathbb{E}_{\mathbf{y} \sim \pi_\theta(\cdot|x)} [r] \right].
\end{equation}
To achieve this objective, various RL algorithms have been adapted for the language modeling context, including the widely used Proximal Policy Optimization (PPO)~\cite{schulman2017proximal} and its variant that forgoes the critic model, Group Relative Policy Optimization (GRPO)~\cite{shao2024deepseekmath}. 
For example, GRPO proposed estimating the baseline from group scores to eliminate the need for a critic model, maximizing the following objective:
\begin{equation}
\begin{split}
    \mathcal{J}_{\text{GRPO}}(\theta) = & \mathbb{E}_{x \sim \mathcal{D}} \bigg[ \frac{1}{G} \sum_{i=1}^G \frac{1}{|\mathbf{y}_i|} \sum_{t=1}^{|\mathbf{y}_i|} \bigg( \min \Big( \rho_{i,t}(\theta) \hat{A}_{i,t}, \\
    & \text{clip}\big(\rho_{i,t}(\theta), 1-\epsilon, 1+\epsilon\big) \hat{A}_{i,t} \Big) - \beta \mathbb{D}_{\text{KL}} \bigg) \bigg],
\end{split}
\label{obj:grpo}
\end{equation}
where $\rho_{i,t}(\theta) = \frac{\pi_\theta(y_{i,t}|x, \mathbf{y}_{i,<t})}{\pi_{\theta_{\text{old}}}(y_{i,t}|x, \mathbf{y}_{i,<t})}$ denotes the probability ratio for the $t$-th token of the $i$-th sampled response $\mathbf{y}_i$ with length $|\mathbf{y}_i|$, and $\epsilon$ is the clipping parameter. The responses form a group of size $G$ sampled from the old policy $\{\mathbf{y}_i\}_{i=1}^G\sim\pi_{\theta_{\text{old}}}(\cdot|x)$.
The advantage $\hat{A}_{i,t} = \frac{r_i - \text{mean}(r)}{\text{std}(r)}$ is computed by normalizing the rewards within the group. The KL divergence term $\mathbb{D}_{\text{KL}}$ (\textit{i.e.}, $\mathbb{D}_{\text{KL}}(\pi_\theta || \pi_{\text{ref}})$) is explicitly added to the loss to regularize the policy towards the reference model $\pi_{\text{ref}}$. {More algorithms are detailed in Appendix}~\ref{app:algorithms}. It is worth noting that these algorithms belong to the policy gradient class of RL~\cite{williams1992simple,sutton1998reinforcement}, which update the model by learning exclusively from on-policy samples, thus maximizing the likelihood of trajectories that yield positive rewards while suppressing those leading to incorrect outcomes.

\subsection{Evaluation Metrics}

The measurement of the \textit{reasoning capability boundary} of a model is challenging and highly dependent on the appropriate metrics.
Standard evaluation metrics, such as greedy decoding accuracy and the average performance of nucleus sampling~\cite{holtzmancurious}, primarily reflect the model's tendency to converge on the most probable output, thus failing to capture the full extent of a model's \text{reasoning capability boundary}. Furthermore, methods such as Best-of-N~\cite{cobbe2021training} and majority voting, which are used for selecting correct answers, may overlook the model's potential ability to generate correct paths. Therefore, recent work has widely adopted Pass@$k$, which focuses exclusively on the \text{existence} of a correct solution within the sampling budget, as the measure of the model's reasoning capability boundary~\cite{yue2025does,zhu2025surprising}. 
In this paper, we used the same unbiased estimation method to calculate Pass@$k$, which eliminates the unstable estimation caused by high variance when directly computing Pass@$k$ with $k$ sampled outputs per problem~\cite{chen2021evaluating}.
Specifically, it first generates $n$ samples ($n\geq k$) for each problem $x$ and counts the number of correct responses as $c$. Then, it computes an unbiased estimate of Pass@$k$ as:
\begin{equation}
    \text{Pass}@k = \mathbb{E}_{x \sim \mathcal{D}} \left[ 1 - \frac{\binom{n-c}{k}}{\binom{n}{k}} \right].
\label{eq:passk}
\end{equation}
The magnitude of $k$ reveals different aspects of the behavior of the model. When $k$ is sufficiently large, Pass@$k$ reflects the model's exploration boundary. 
When $k=1$, Pass@$1$ approximates greedy decoding accuracy, which serves as a proxy for the reliability of the model's dominant path.

\subsection{Empirical Insights from Recent Work}

Recent work has challenged common understanding of the mechanism of RLVR. It reveals that instead of continuously generating and reinforcing new correct paths as in traditional RL~\cite{mnih2015human,silver2017mastering,sallab2017deep}, RLVR enhances the reasoning capabilities of LLMs by amplifying the probability distribution of existing correct paths in the base model~\cite{yue2025does,zhu2025surprising}.
Specifically, common RL algorithms used for RLVR (\textit{e.g.} PPO, GRPO and REINFORCE) tend to oversharpen the greedy decoding distribution towards the correct paths through learning. Although these algorithms effectively improve the performance of greedy decoding (Pass@$1$ accuracy), it comes at the cost of compromising the model's reasoning capability boundaries. After RLVR training, the learned model's Pass@$k$ performance is even worse than the base model, \textit{i.e.}, the learned model loses the capacity to generate correct paths for some problems that could be correctly solved by the base model.

These works have theoretically and empirically dissected the mechanisms driving this reasoning boundary shrinkage induced by RLVR. 
Yue et al.~\cite{yue2025does} characterize this process as an \textit{identify-and-amplify} mechanism, showing that standard RL algorithms improve Pass@$1$ by aggressively shifting probability mass to high-likelihood correct paths, thus shrinking the diverse solution space latent in the base model. 
Furthermore, Zhu et al.~\cite{zhu2025surprising} provide a rigorous gradient analysis by decomposing the RLVR objective into two distinct sub-objectives, Positive Sample Reinforcement (PSR) and Negative Sample Reinforcement (NSR). Therefore, the RLVR objective optimizes the expected reward-weighted likelihood as:
\begin{equation}
\begin{split}
    \mathcal{J}(\theta) & = \mathbb{E}_{x \sim \mathcal{D}} \left[ \sum_{\mathbf{y}} r \cdot \pi_\theta(\mathbf{y}|x) \right] , \quad r \in \{-1, +1\} \\
    & = \underbrace{\mathbb{E}_{x \sim \mathcal{D}} \left[ \sum_{\mathbf{y} \in \mathcal{Y}^+} \pi_\theta(\mathbf{y}|x) \right]}_{\mathcal{J}_{\text{PSR}}(\theta)} - \underbrace{\mathbb{E}_{x \sim \mathcal{D}} \left[ \sum_{\mathbf{y} \in \mathcal{Y}^-} \pi_\theta(\mathbf{y}|x) \right]}_{\mathcal{J}_{\text{NSR}}(\theta)},
\end{split}
\end{equation}
where $\mathcal{Y}^+$ and $\mathcal{Y}^-$ denote the sets of correct and incorrect responses, respectively. 
The PSR component is similar to supervised fine-tuning (SFT), updating the model to increase the likelihood of correct responses. It mathematically necessitates suppressing the probabilities of all other paths, including potentially correct but unexplored paths, thereby explicitly reducing policy entropy and causing the reasoning boundary to contract. 
In contrast, the NSR component shows the unique advantage in RLVR. By suppressing incorrect paths, it \textit{implicitly redistributes} the probability mass among the remaining paths guided by the model's prior knowledge, improving the full Pass@$k$ spectrum.
Crucially, this behavior differs fundamentally from entropy regularization, which indiscriminately flattens the output distribution. As analyzed by Zhu et al.~\cite{zhu2025surprising}, the NSR gradient is dampened by a factor of $(1-\pi)$, which reduces the penalty magnitude for high-confidence tokens. This mechanism prevents destructive updates to the model's valid priors (\textit{e.g.}, linguistic competence) embedded within incorrect paths, thus strictly preserving the base model's intrinsic capabilities while correcting reasoning errors.

However, RLVR training that is fully based on the NSR component faces limitations. NSR fails to provide the contrastive baseline required to reduce gradient variance, leading to slow convergence in exploitation (Pass@$1$) and potential training instability~\cite{shao2024deepseekmath,yu2025dapo}.
Furthermore, it operates primarily as a \textit{pruning} mechanism, which requires the base model to already have a strong prior to implicitly infer correct paths via elimination.
In scenarios where the solution space is vast or the model's initial capability is weak (\textit{i.e.}, cold start phase), pure NSR lacks the signal necessary to establish correct reasoning patterns~\cite{zhu2025surprising,wang2025icpo}. 
Therefore, since RLVR works by amplifying the probability of the correct paths rather than generating new correct paths, its core challenge lies in improving the efficiency of sampling the correct paths (Pass@$1$) while protecting the reasoning boundary (Pass@$k$). Methodologically, this requires the RL algorithms to be well-designed and effectively balance the advantage magnitude of PSR and NSR.

\section{Methodology}
\label{sec:method}

In this section, we propose Asymmetric Group Policy Optimization (AGPO), a NSR-dominated dynamic advantage estimation method designed to address the trade-off between sampling efficiency (Pass@$1$) and reasoning capability boundary preservation (Pass@$k$) in RLVR.

\subsection{Motivation: The Dilemma of Unconstrained PSR}

Unconstrained amplification of the probability of sampled correct paths, typical in standard PSR, often causes the output distribution of the trained model to excessively sharpen towards frequently occurring correct paths.
This bias is intrinsic to the \textit{on-policy} sampling mechanism of RLVR, where frequently occurring correct paths are sampled disproportionately often, thereby accumulating the majority of positive reinforcement. In contrast, long-tail correct paths are sampled sparsely, receiving minimal positive reinforcement signals, which further suppresses their generation probability in the optimized policy.
To formalize this, given a specific prompt $x$, the \textit{prior correctness probability} $\mathcal{P}_{\Theta}(x)$ of the base model $\Theta$ is defined as
\begin{equation}
    \mathcal{P}_{\Theta}(x) = \mathbb{E}_{\mathbf{y} \sim \pi_{\Theta}(\cdot|x)} [\mathbb{I}(\mathbf{y} \in \mathcal{Y}^+)],
\end{equation}
where $\pi_{\Theta}$ represents the pre-trained base policy, $\mathcal{Y}^+$ denotes the set of correct response sequences, and $\mathbb{I}(\cdot)$ is the indicator function. 
The probability $\mathcal{P}_{\Theta}(x)$ quantifies the base model's intrinsic capability on a specific instance $x$.
We analyze the optimization dynamics across three distinct regimes of $\mathcal{P}_{\Theta}(x)$:
\begin{itemize}
    \item $\mathcal{P}_{\Theta}(x) \to 0$. The prompt exceeds the model's current capabilities. In this regime, RLVR inherently struggles to generate fundamentally new correct paths without external reinforcement signal~\cite{yue2025does}.
    \item $\mathcal{P}_{\Theta}(x) \to 1$. The model has already mastered the problem latent in the prompt, with the output entropy typically being low. Continued reinforcement in this regime yields marginal improvement for the reasoning capability. Worse, it aggravates the collapse of the output distribution onto these low-entropy trivial paths, which limits exploration and degrades Pass@$k$.
    \item $0 < \mathcal{P}_{\Theta}(x) < 1$. This is the fundamental of RLVR. Although the correct paths already exist in the output distribution of the base model, the model cannot confidently output the correct path, with the output entropy typically being high. 
    In this regime, RLVR utilizes contrastive signals to distinguish correct paths from incorrect ones, thus converting the model's potential capacity into better performance.
\end{itemize}
Under the on-policy sampling paradigm in RLVR, the reward accumulation for the correct paths is proportional to $\mathcal{P}_{\Theta}(x)$. 
Applying static positive rewards for any sampled correct path inevitably biases optimization towards already dominant modes, accelerating output distribution collapse. This drives a dynamic reinforcement strategy, where the advantage magnitude of the positive reward should be inversely correlated with the sampling frequency, reducing the reward of trivial paths while amplifying the encouragement for rare, correct paths.

\subsection{Asymmetric Group Policy Optimization}

We empirically estimate $\mathcal{P}_{\Theta}(x)$ by sampling a group of responses $\{{\mathbf{y}_i\}_{i=1}^G \sim \pi_{\theta_{\text{old}}}}$ for a given prompt $x$. Consequently, AGPO adopts a group-based objective framework similar to GRPO shown in Eq.~(\ref{obj:grpo}), formulated as
\begin{equation}
\begin{split}
    \mathcal{J}_{\text{AGPO}}(\theta) = &\mathbb{E}_{x \sim \mathcal{D}} \bigg[ \frac{1}{G} \sum_{i=1}^G \frac{1}{|\mathbf{y}_i|} \sum_{t=1}^{|\mathbf{y}_i|} \min \Big( \rho_{i,t}(\theta) \hat{A}_i^{\text{AGPO}}, \\
    & \text{clip}(\rho_{i,t}(\theta), 1-\epsilon, 1+\epsilon) \hat{A}_i^{\text{AGPO}} \Big) - \beta \mathbb{D}_{\text{KL}} \bigg) \bigg],
\end{split}
\end{equation}
where $\rho_{i,t}(\theta)$ is the token-level probability ratio defined as in Eq.~(\ref{obj:grpo}), and $\epsilon$ is the clipping parameter. $\hat{A}_i^{\text{AGPO}}$ denotes the advantage for the response $\mathbf{y}_i$, where the subscript $t$ is omitted as the outcome-based reward applies uniformly to the entire sequence. The core of AGPO lies in the design of the group advantage function $\hat{A}_i^{\text{AGPO}}$, which introduces an asymmetric mechanism to decouple the update dynamics of positive and negative samples, defined as
\begin{equation}
    \hat{A}_i^{\text{AGPO}} = \underbrace{  \frac{1}{\sqrt{\sigma^2 + \delta^2}} \cdot (r_i - \mu) }_{\text{Constrained group relative term}} + \underbrace{\mathbb{I}(r_i < 0) \cdot \mathcal{R}}_{\text{Gated negative term}},
\end{equation}
where $r_i\in\{-1,+1\}$ is the reward for the response $\mathbf{y}_i$, $\mu$ and $\sigma$ denote the mean and standard deviation of the group rewards, and $\delta$ serves as the constraint factor. $\mathcal{R}<0$ is used to guarantee a basic negative reinforcement magnitude for incorrect paths. Following the empirical setting of the NSR component in W-REINFORCE~\cite{zhu2025surprising}, we set $\mathcal{R}$ to $-1$ by default in our experiments.

$\hat{A}_i^{\text{AGPO}}$ differs from the standard advantage function of GRPO in two aspects. 
First, in the \textit{constrained group relative term}, $\delta$ acts as a variance regularizer, which constrains the advantage magnitude, thereby preventing sharpening of the output distribution.
Second, the \textit{gated negative term} decouples the negative reinforcement for incorrect paths from group statistics. By enforcing a stable negative signal $\mathcal{R}$, it ensures consistent \text{pruning} of the search space even when all responses are incorrect. Furthermore, it ensures that the overall advantage magnitude function $\hat{A}_i^{\text{AGPO}}$ is dominated by NSR, in order to maintain the reasoning capability boundary as much as possible in RLVR training.
Notably, due to the NSR-dominate advantage strategy similar to W-REINFORCE, AGPO shows reduced dependency on KL regularization~\cite{zhu2025surprising}. Our empirical results demonstrate that while a minimal KL constraint guarantees training stability, completely discarding the KL constraint can, in some cases, yield superior performance.

Figure~\ref{fig:reward_dynamics} illustrates the comparison of the advantage dynamics among AGPO, GRPO, and W-REINFORCE. 
Overall, AGPO shows a nearly linear trend, avoiding the sharp spikes observed in GRPO as the sample distribution shifts.
In the NSR regime (Right), AGPO imposes a consistent negative reinforcement (starting at -1) to suppress incorrect paths.
In addition, compared to W-REINFORCE which applies a \text{static} reward even in trivial samples (all correct paths), AGPO is dynamic, in that the parameter $\delta$ regulates the relative advantage magnitude. As the value of $\delta$ increases, the difference in relative advantage decreases.

\begin{table*}[ht]
\centering
\caption{Comparison of Pass@$k$ (\%) results on \texttt{MATH}, \texttt{Olympiad} and \texttt{AIME-2024} with Qwen2.5-Math-7B. \textbf{Bold} and \underline{underlined} numbers denote the best and second-best results, respectively.}
\label{tab:main_results}
\begin{tabularx}{0.92\textwidth}{cl YYYYYYYYY}
\toprule
\multirow{2}{*}{\textbf{Dataset}} & \multirow{2}{*}{\textbf{Model}} & \multicolumn{9}{c}{\textbf{Pass@$k$}} \\
\cmidrule(l){3-11}
 & & $1$ & $2$ & $4$ & $8$ & $16$ & $32$ & $64$ & $128$ & $256$ \\
\midrule

% % -------- MATH --------
% \multirow{8}{*}{\rotatebox[origin=c]{90}{\texttt{MATH}}} 
%  & Base Model & $63.2$ & $76.0$ & 83.7 & 88.4 & 91.6 & 93.7 & 95.2 & 96.2 & 96.9 \\
%  & PPO & 76.6 & 82.6 & 86.7 & 89.6 & 91.7 & 93.4 & 94.7 & 95.6 & 96.3 \\
%  & PPO-version2 & 75.8 & 81.5 & 85.5 & 88.4 & 90.6 & 92.2 & 93.4 & 94.3 & 94.8 \\
%  & GRPO & 76.3 & 81.7 & 85.6 & 88.4 & 90.6 & 92.3 & 93.6 & 94.7 & 95.5 \\
%  & GRPO-version2 & 76.2 & 81.5 & 85.2 & 88.0 & 90.0 & 91.7 & 93.0 & 94.0 & 94.7 \\
%  & REINFORCE & 74.1 & 78.7 & 81.9 & 84.4 & 86.8 & 88.2 & 89.6 & 91.0 & 92.3 \\
%  & W-REINFORCE & 76.3 & 82.4 & 86.9 & 90.2 & 92.3 & 93.8 & 95.4 & 96.2 & 96.6 \\
%  & \cellcolor{rowblue}\textbf{AGPO} & \cellcolor{rowblue} 77.0 & \cellcolor{rowblue} 83.2 & \cellcolor{rowblue} 87.6 & \cellcolor{rowblue} 90.5 & \cellcolor{rowblue} 92.6 & \cellcolor{rowblue} 94.1 & \cellcolor{rowblue} 95.2 & \cellcolor{rowblue} 95.8 & \cellcolor{rowblue} 96.4 \\
% \midrule
% -------- MATH --------
\multirow{6}{*}{\rotatebox[origin=c]{90}{\texttt{MATH}}} 
 & Base Model & 63.1 & 76.2 & 83.5 & 88.1 & 91.5 & 93.6 &{95.3} & \underline{96.2} & \textbf{96.7} \\
% & Base Model & 62.2 & 75.2 & 82.8 & 87.2 & 90.5 & 92.8 & {94.6} & {95.4} & \underline{96.1} \\
 % & Base Model-version2 & 54.3 & 69.1 & 78.7 & 84.5 & 88.0 & 90.4 & \underline{92.0} & \textbf{93.1} & \textbf{94.0} \\
 % & PPO (Old) & 76.6 & 82.6 & 86.7 & 89.6 & 91.7 & 93.4 & 94.7 & 95.6 & 96.3 \\
 & PPO & 75.8 & 81.5 & 85.5 & 88.4 & 90.6 & 92.2 & 93.4 & 94.3 & 94.8 \\
 % & GRPO (Old) & 76.3 & 81.7 & 85.6 & 88.4 & 90.6 & 92.3 & 93.6 & 94.7 & 95.5 \\
 & GRPO & 76.2 & 81.5 & 85.2 & 88.0 & 90.0 & 91.7 & 93.0 & 94.0 & 94.7 \\
 & REINFORCE & 74.1 & 78.7 & 81.9 & 84.4 & 86.8 & 88.2 & 89.6 & 91.0 & 92.3 \\
 & W-REINFORCE & \underline{76.3} & \underline{82.4} & \underline{86.9} & \underline{90.2} & \underline{92.3} & \underline{93.8} & \underline{95.4} & \underline{96.2} & \underline{96.6} \\
 & \cellcolor{rowblue}\text{AGPO} & \cellcolor{rowblue} \textbf{77.2} & \cellcolor{rowblue} \textbf{83.5} & \cellcolor{rowblue} \textbf{87.9} & \cellcolor{rowblue} \textbf{90.7} & \cellcolor{rowblue} \textbf{92.7} & \cellcolor{rowblue} \textbf{94.3} & \cellcolor{rowblue} \textbf{95.6} & \cellcolor{rowblue} \textbf{96.4} & \cellcolor{rowblue} \underline{96.6} \\
\midrule

% -------- Olympiad --------
\multirow{6}{*}{\rotatebox[origin=c]{90}{\texttt{Olympiad}}} 
 & Base Model & 25.0 & 35.2 & 44.4 & 52.0 & 58.3 & 63.4 & {67.6} & {71.3} & {74.4} \\
 & PPO & 37.1 & 43.7 & 49.8 & 55.1 & 59.6 & 63.2 & 66.3 & 69.1 & 71.9 \\
 & GRPO & \underline{37.5} & \underline{44.2} & 50.3 & 55.7 & 60.2 & 63.8 & 66.9 & 69.5 & 72.0 \\
 & REINFORCE & 35.1 & 42.5 & 47.9 & 52.3 & 57.8 & 61.4 & 66.7 & 69.3 & 72.5 \\
 & W-REINFORCE & 36.9 & \underline{44.2} & \underline{50.8} & \underline{56.5} & \underline{61.3} & \underline{65.4} & \underline{69.0} & \underline{72.1} & \underline{75.1} \\
 & \cellcolor{rowblue}\text{AGPO} & \cellcolor{rowblue}\textbf{38.6} & \cellcolor{rowblue}\textbf{45.4} & \cellcolor{rowblue}\textbf{51.7} & \cellcolor{rowblue}\textbf{57.5} & \cellcolor{rowblue}\textbf{62.0} & \cellcolor{rowblue}\textbf{65.9} & \cellcolor{rowblue}\textbf{69.4} & \cellcolor{rowblue}\textbf{72.5} & \cellcolor{rowblue}\textbf{75.3} \\
\midrule

% % -------- AIME-2025 --------
% \multirow{6}{*}{\rotatebox[origin=c]{90}{\texttt{AIME-2025}}} 
% % & Base Model & 6.1 & 9.7 & 13.8 & 17.9 & 22.2 & 26.5 & 30.8 & 36.6 & 46.7 \\
%  & Base Model & 6.5 & 10.5 & 15.1 & 19.8 & 24.7 & 29.2 & 33.3 & 37.5 & 43.3 \\
% % & PPO & 8.5 & 13.2 & 18.0 & 22.5 & 26.6 & 30.3 & 33.8 & 37.9 & 43.3 \\
%  & PPO & 9.6 & 14.5 & 19.7 & 24.7 & \underline{30.0} & \underline{35.5} & \underline{41.2} & 47.1 & 53.3 \\
% % & GRPO & 10.3 & 14.7 & 19.4 & 24.0 & 28.4 & 32.8 & 37.3 & 42.5 & 50.0 \\
%  & GRPO & 10.1 & 14.4 & 19.1 & 24.2 & 29.6 & 35.0 & 40.0 & 44.9 & 50.0 \\
%  & REINFORCE & 9.1 & 12.3 & 16.0 & 20.9 & 26.3 & 32.3 & 38.7 & 45.0 & 50.3 \\
%  & W-REINFORCE & \underline{10.5} & \underline{15.4} & \underline{20.2} & \underline{24.8} & 29.7 & 34.4 & 40.8 & \underline{47.9} & \underline{56.5} \\
%  & \cellcolor{rowblue}\textbf{AGPO} & \cellcolor{rowblue}\textbf{11.2} & \cellcolor{rowblue}\textbf{15.9} & \cellcolor{rowblue}\textbf{21.3} & \cellcolor{rowblue}\textbf{26.6} & \cellcolor{rowblue}\textbf{32.3} & \cellcolor{rowblue}\textbf{37.9} & \cellcolor{rowblue}\textbf{43.6} & \cellcolor{rowblue}\textbf{50.6} & \cellcolor{rowblue}\textbf{57.1} \\
% \bottomrule

% -------- AIME-2024 --------
\multirow{6}{*}{\rotatebox[origin=c]{90}{\texttt{AIME-2024}}} 
 & Base Model & 13.9 & 21.8 & 29.9 & 36.1 & 41.5 & 47.4 & 54.2 & 61.7 & 70.0 \\
 & PPO & 25.4 & 32.8 & 38.7 & 44.8 & 50.9 & 55.9 & 59.9 & 64.9 & 73.3 \\
 & GRPO & \underline{28.4} & 35.7 & 41.9 & 47.5 & 52.8 & 57.4 & 61.7 & 67.0 & 73.3 \\
 & REINFORCE & 24.7 & 33.9 & 40.5 & 47.0 & 52.9 & 58.2 & 62.1 & 66.2 & 71.9 \\
 & W-REINFORCE & 27.9 & \underline{36.0} & \underline{42.7} & \underline{49.0} & \underline{55.1} & \underline{60.1} & \underline{65.0} & \underline{69.8} & \underline{73.5} \\
 & \cellcolor{rowblue}\text{AGPO} & \cellcolor{rowblue} \textbf{30.1} & \cellcolor{rowblue} \textbf{39.0} & \cellcolor{rowblue} \textbf{45.8} & \cellcolor{rowblue} \textbf{51.1} & \cellcolor{rowblue} \textbf{55.8} & \cellcolor{rowblue} \textbf{60.3} & \cellcolor{rowblue} \textbf{65.2} & \cellcolor{rowblue} \textbf{71.3} & \cellcolor{rowblue} \textbf{76.8} \\
\bottomrule
\label{tab:qwenmath27}
\end{tabularx}
\end{table*}

\begin{figure*}[ht]
    \centering
    \includegraphics[width=0.98\textwidth]{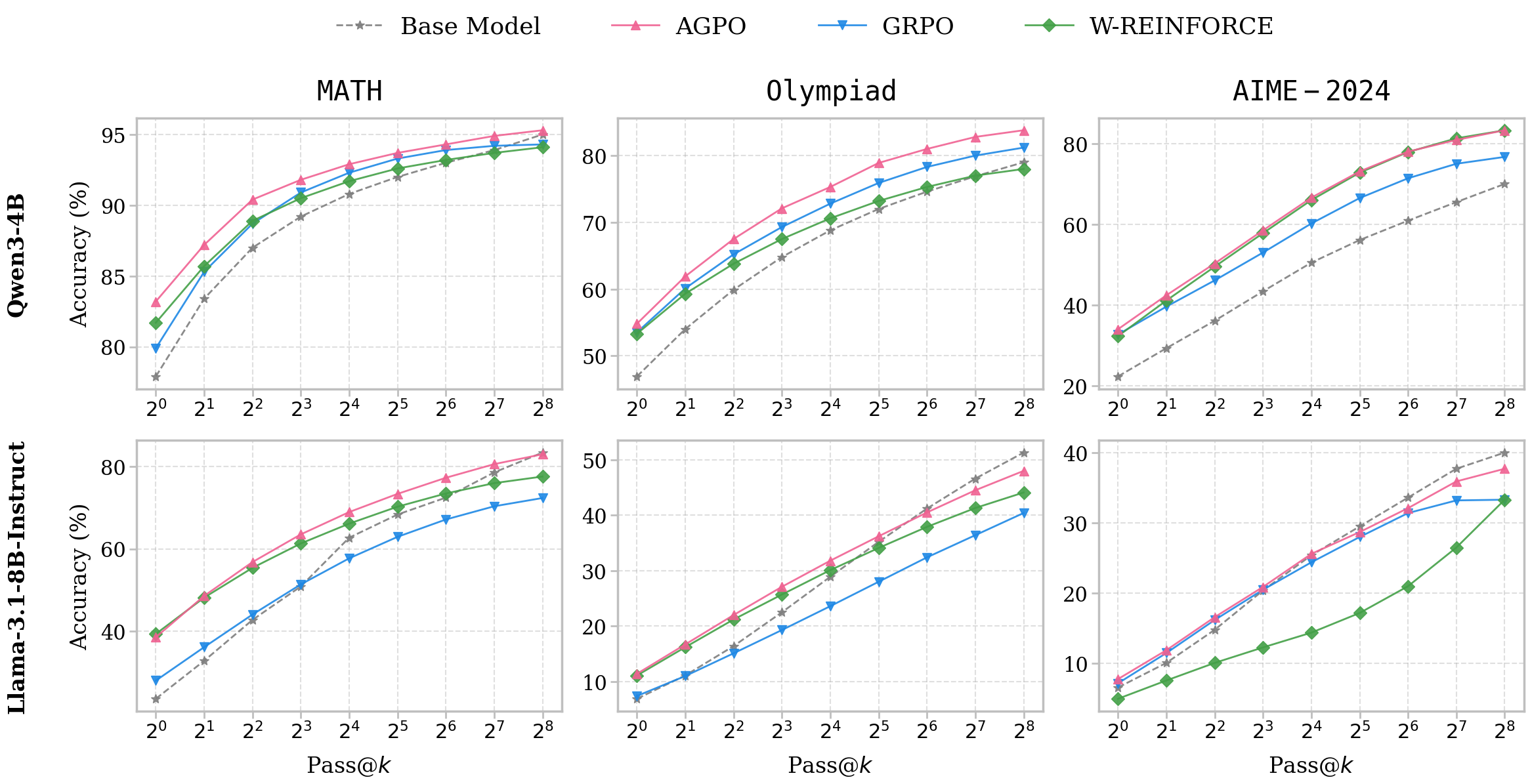}
    \caption{Pass@$k$ performance scaling on \texttt{MATH}, \texttt{Olympiad} and \texttt{AIME-2024} with Qwen3-4B and Llama-3.1-8B-Instruct.
    }
    \label{fig:pass_k_results}
\end{figure*}

\begin{figure*}[ht]
    \centering
    \includegraphics[width=0.99\textwidth]{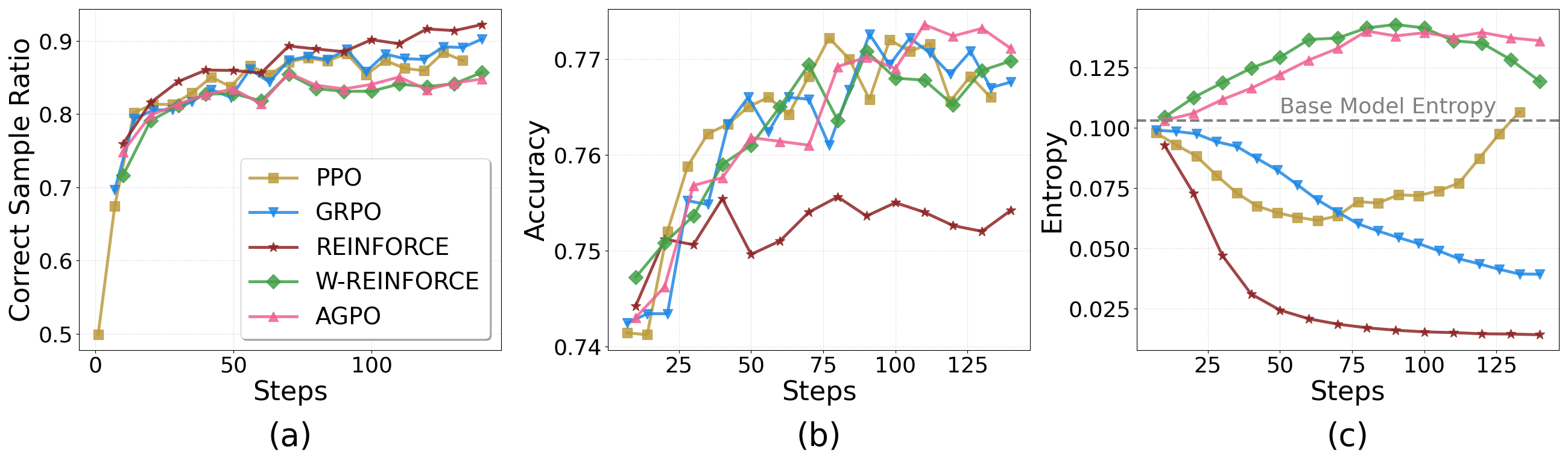}
    \caption{Training dynamics of Qwen2.5-Math-7B on \texttt{MATH} under different RL methods across training steps. (a) Correct responses ratio per batch on the training set. (b) Greedy decoding accuracy (Pass@$1$) on the test set. (c) The model’s entropy on the test set.
    }
    \label{fig:training_dynamics}
\end{figure*}

\section{Experiments on Mathematical Reasoning}
\label{sec:experiments}

\subsection{Experimental Setup}
\textbf{Models and Benchmarks.} 
To ensure the generality of our conclusions, we conduct experiments across multiple LLM families, including Qwen2.5-Math-7B~\cite{yang2024qwen2}, Qwen3-4B~\cite{yang2025qwen3}, and Llama-3.1-8B-Instruct~\cite{grattafiori2024llama}.
Notably, we use Qwen3-4B with the non-thinking mode for training and inference.
In the RLVR training phase, we use the standard \texttt{MATH} dataset~\cite{hendrycks2021measuring}, containing 7,500 problems.
We evaluate on five widely used mathematical reasoning benchmarks, including \texttt{MATH}~\cite{hendrycks2021measuring}, \texttt{Olympiad}~\cite{he2024olympiadbench}, \texttt{AIME-2024}~\cite{shao2024deepseekmath}, \texttt{AIME-2025}~\cite{yang2025qwen3}, and \texttt{GaoKao-2024}~\cite{liu2025matheval}.
These benchmarks cover a diverse spectrum of difficulty and formats. For example, \texttt{MATH} serves as a comprehensive collection of general mathematical problems. \texttt{Olympiad}, \texttt{AIME-2024}, and \texttt{AIME-2025} are highly challenging competition-level problems requiring deep reasoning. \texttt{GaoKao-2024} is a set of standardized multiple-choice problems derived from the Chinese college entrance examination.
Due to space constraints, the detailed experimental results for \texttt{AIME-2025} and \texttt{GaoKao-2024} are reported in the Appendix~\ref{app:morere}.

\textbf{Compared Algorithms and Evaluation Setup.}
We compare the performance of AGPO with base model and commonly used RL algorithms, including PPO~\cite{schulman2017proximal}, GRPO~\cite{shao2024deepseekmath}, REINFORCE~\cite{williams1992simple}, and W-REINFORCE~\cite{zhu2025surprising}. We train the models using VeRL~\cite{sheng2025hybridflow}, a popular open-source framework that offers robust and reproducible implementations of standard RL algorithms for LLMs~\cite{yue2025does}. Following recent work~\cite{zhu2025surprising},
during evaluation, we sample 256 responses per problem and calculate Pass@$k$ metrics using Eq.~(\ref{eq:passk}). Furthermore, details of the experimental parameters and the prompt templates used are provided in the Appendices~\ref{app:expset}, respectively.

\subsection{Main Pass@$k$ Results}

Table~\ref{tab:qwenmath27} shows the Pass@$k$ results of various RL methods with Qwen2.5-Math-7B on three mathematical benchmarks. It can be seen that AGPO achieves the best accuracy (Pass@$1$) across all reasoning benchmarks while effectively counteracting the reasoning boundary shrinkage commonly observed in standard RLVR algorithms ($k>1$). 
For example, on the challenging \texttt{AIME-2024} dataset, AGPO improves the accuracy of Qwen2.5-Math-7B from 13.9\% to 30.1\%, surpassing all baselines. More importantly, while standard algorithms such as PPO and GRPO show performance degradation compared to the base model at $k=256$ (\textit{e.g.}, dropping from 96.7\% to 94.8\% and 94.7\% respectively on \texttt{MATH} dataset), AGPO maintains a coverage of 96.6\%, showing the ability to preserve the base model's reasoning boundary.
It should be noted that when $k=256$, AGPO outperforms the base model on both the \texttt{Olympiad} and \texttt{AIME-2024} datasets. This is because on these specific datasets, the $k$ value often needs to be large enough for the base model to outperform the RL methods~\cite{yue2025does}.

The inference scaling trends shown in Figure~\ref{fig:pass_k_results} further illustrate the robustness of AGPO across different base models. Overall, on both Qwen3-4B and Llama-3.1-8B-Instruct models, AGPO achieved the best inference performance improvement compared to GRPO and W-REINFORCE.
Notably, for the Llama-3.1-8B-Instruct model, which shows pronounced boundary shrinkage after RL training, AGPO demonstrates the closest Pass@$256$ results to the base model.
These results demonstrate that AGPO, by employing an NSR-dominated dynamic strategy, suppresses incorrect inference paths while simultaneously enhancing rare correct paths, ensuring that the model improves its exploitability while reducing inference boundary shrinkage.

\subsection{Training Dynamics Analysis}

The training dynamics illustrated in Figure~\ref{fig:training_dynamics} provide quantitative insights into how AGPO balances exploitation efficiency with the preservation of the reasoning boundary.

Figure~\ref{fig:training_dynamics}(b) illustrates the generalization accuracy of the RL methods. Notably, AGPO shows a more gradual initial increase in test set accuracy compared to PPO and GRPO. This is because PPO and GRPO reinforce correct paths with greater magnitude, allowing them to rapidly exploit a subset of high-likelihood paths. However, these methods eventually hit a performance bottleneck as their exploration capacity diminishes. In contrast, AGPO achieves a slight lead in later stages. Although its initial learning is slower, AGPO’s ability to maintain high entropy results in more robust and continuous optimization, effectively mining the latent reasoning patterns present in the base model. 

The critical role of entropy preservation is shown in Figure~\ref{fig:training_dynamics}(c). NSR-dominated methods, including AGPO and W-REINFORCE, demonstrate superior entropy maintenance. Unlike PPO, GRPO, and REINFORCE, whose entropy levels quickly collapse below the base model baseline (dashed line), AGPO and W-REINFORCE maintain levels higher than the base model. This explains the importance of NSR-dominated strategies in RLVR, which maintain the model's self-exploration capabilities during training while preserving the model's inference boundaries by maintaining the model's output entropy.
Compared to W-REINFORCE, APGO's advantage lies in its use of dynamic PSR, which avoids continuously reinforcing trivial correct paths. This allows AGPO to maintain higher output entropy in the later stages of training and achieve greater accuracy improvement on the test set.
Notably, REINFORCE achieved the best fit on the training data shown in Figure~\ref{fig:training_dynamics}(a), but the entropy value dropped sharply to near zero, where the distribution tends to concentrate on a narrow set of paths.
This highlights the importance of maintaining the model's output entropy for paths exploration during the RLVR training.

\section{Application: Search Ads Relevance}

\subsection{Search Ads Relevance Optimization}

In large-scale \textit{e-commerce search advertising systems}, such as JD Ads Search~\cite{fan2019mobius,liu2021que2search}, the relevance model serves as the critical \text{gatekeeper} that filters out semantically mismatched advertisements before entering the ranking stage~\cite{lu2020twinbert}.
This module operates downstream of candidate retrieval, ensuring that only semantically aligned advertisements proceed to the ranking stage.
Formally, given a user query $q$ and a candidate product $p$, the relevance model functions as a classifier to denote the relevance level.
Due to strict latency constraints in online serving, the industry standard adopts a teacher-student distillation paradigm~\cite{liang2023less,jiao2025taosr}.
In this pipeline, a powerful LLM acts as the \text{teacher} to generate high-quality relevance labels, which are then used to train a lightweight \text{student} model (\textit{e.g.}, BERT-based encoder) for online inference. Therefore, the performance of the entire relevance module is upper-bounded by the labeling accuracy of the LLM teacher.
Optimizing the teacher model typically relies on human-annotated labels as ground truth.
Since human annotations provide only the relevance label without explicit CoT reasoning, employing RL training is often more effective than direct SFT~\cite{jiao2025taosr}.
At this point, optimizing the teacher model follows a standard RLVR paradigm, typically employing methods such as PPO and GRPO.
However, its training faces the same dilemma in RLVR, where the optimization objective often leads the policy to bias towards frequently occurring $(q,p)$ pairs and simple semantic cases, and the model's ability to distinguish between long-tail $(q,p)$ pairs and complex cases decreases~\cite{fan2019mobius,liu2021que2search}.

\subsection{Experiments on Search Ads Relevance Reasoning}

\textbf{Experimental Setup.} 
We use Rele-Ads-8B, a decoder-only transformer model , which serves as the core teacher model in the production pipeline of JD Ads Search, one of the largest e-commerce search advertising platforms serving hundreds of millions of active users. This model is used to generate high-quality labeling data for training a lightweight online student model, \textit{i.e.}, a three-layer BERT model.
Rele-Ads-8B follows a three-tier classification schema: \textit{irrelevant}, \textit{partially relevant}, and \textit{fully relevant}, where it explicitly generates a CoT reasoning path before generating the inference class.
To optimize Rele-Ads-8B, we use a real-world e-commerce relevance dataset sourced from JD's production logs.
It contains 90,000 training samples and 20,000 validation samples, where each entry consists of a user query, a candidate product, and a ground truth (GT) label. The GT label for each data point is cross-validated by multiple annotators.
More experimental details are provided in the Appendix~\ref{app:expset}.

\textbf{Evaluation Metrics.} 
In e-commerce search advertising systems, the relevance module is responsible for both user experience and platform revenue, which means it needs to filter out irrelevant advertising products as much as possible while retaining relevant ones~\cite{agrawal2023enhancing,jiao2025taosr}. Therefore, the effectiveness of the model is measured by the proportion of irrelevant products (PIR) in the products presented to users and online revenue metrics~\cite{cheng2016wide,zhou2018deep}. These metrics are formalized in detail in Appendix~\ref{app:eva}.

\begin{table}[t]
\centering
\caption{Comparison of PIR (\%) and Pass@$k$ (\%) results on search ads relevance reasoning with Rele-Ads-8B.}
\label{tab:model_comparison}
\newcolumntype{C}{>{\centering\arraybackslash}X}
\begin{tabularx}{\linewidth}{lCCCC}
\toprule
\textbf{Model} & \textbf{PIR~($\downarrow$)} & \textbf{Pass@1} & \textbf{Pass@2} & \textbf{Pass@4} \\ 
\midrule
Base Model     & 7.2      & 78.0  &  81.9  & \underline{84.2} \\
GRPO           & \underline{5.8}       & \underline{80.2}  &  \underline{82.4}  & 83.7  \\
\cellcolor{rowblue}AGPO           &\cellcolor{rowblue}\textbf{5.4}       & \cellcolor{rowblue}\textbf{80.5}  &  \cellcolor{rowblue}\textbf{83.1}  & \cellcolor{rowblue}\textbf{85.0}   \\ 
\bottomrule
\end{tabularx}
\label{tab:teacher_results}
\end{table}

\begin{table}[t]
\centering
\caption{Online A/B test results relative to the online baseline.}
\label{tab:online_ab_minimal}
\resizebox{0.96\linewidth}{!}{ 
    \begin{tabular}{lccc}
    \toprule
     & \textbf{CTRPI} & \textbf{CPM} & \textbf{GMV} \\ 
    \midrule
    \cellcolor{rowblue}$\text{BERT}_\textbf{+AGPO Labels}$  & \cellcolor{rowblue}\textbf{+0.22\%} & \cellcolor{rowblue}\textbf{+0.50\%} & \cellcolor{rowblue}\textbf{+0.21\%} \\ 
    \bottomrule
    \end{tabular}
}
\end{table}

\textbf{Results.}
The offline evaluation of the teacher model is shown in Table~\ref{tab:teacher_results}. The results show that RLVR training effectively decreases semantically mismatched advertisements, with both AGPO and GRPO achieving a significantly lower PIR (the proportion of predicted \textit{fully relevant} cases that are human-annotated as \textit{irrelevant}) than the base model.
Notably, AGPO shows superior performance in Pass@$k$ metrics, demonstrating its capability to maintain a broader reasoning boundary for complex and long-tail query-product pairs. 
These improvements lead to substantial online gains in the production system shown in Table~\ref{tab:online_ab_minimal}. 
The results from a two-day online A/B test reveal gains across core business metrics.
The CTRPI (Click-Through Rate Per Impression) increased by 0.22\%, reflecting improved semantic matching quality. This directly led to a 0.50\% growth in CPM (Cost Per Mille), enhancing the platform's monetization efficiency. Furthermore, a 0.21\% increase in GMV (Gross Merchandise Volume) confirms that the relevance improvements are achieved while maintaining transaction stability.

\section{Conclusion}
\label{sec:conclusion}

We propose AGPO, a new RL method designed to address the reasoning boundary shrinkage problem in RLVR. By employing group-variance-based dynamic advantage estimation and a negative-dominant strategy for incorrect paths, AGPO not only achieves SOTA accuracy on mathematical benchmarks but also demonstrates superior robustness at large sample sizes (\textit{e.g.}, Pass@$256$), effectively preserving the base model's reasoning capability. Furthermore, its deployment in a large-scale search advertising system validates its practical utility in handling real-world semantic matching. 
Future research could explore extending asymmetric strategies to multi-turn dialogue reasoning and investigating more complex dense reward signal scenarios.

% \section*{Impact Statement}

% This paper presents work whose goal is to advance the field of Machine Learning, specifically in the reasoning capabilities of LLMs. The proposed method improves the efficiency and reliability of model training. Potential societal consequences include the development of more capable AI systems, which necessitates responsible deployment and monitoring. We do not feel there are specific negative ethical impacts that must be highlighted here beyond general AI safety concerns.

\bibliography{example_paper} % Remember to create a .bib file
\bibliographystyle{icml2026}

%%%%%%%%%%%%%%%%%%%%%%%%%%%%%%%%%%%%%%%%%%%%%%%%%%%%%%%%%%%%%%%%%%%%%%%%%%%%%%%
%%%%%%%%%%%%%%%%%%%%%%%%%%%%%%%%%%%%%%%%%%%%%%%%%%%%%%%%%%%%%%%%%%%%%%%%%%%%%%%
% APPENDIX
%%%%%%%%%%%%%%%%%%%%%%%%%%%%%%%%%%%%%%%%%%%%%%%%%%%%%%%%%%%%%%%%%%%%%%%%%%%%%%%
%%%%%%%%%%%%%%%%%%%%%%%%%%%%%%%%%%%%%%%%%%%%%%%%%%%%%%%%%%%%%%%%%%%%%%%%%%%%%%%
\newpage
\appendix

% --- 附录标号初始化开始 ---
\setcounter{table}{0}         % 重置表格计数器
\renewcommand{\thetable}{A.\arabic{table}} % 设置表格编号格式为 A.1, A.2...

\setcounter{figure}{0}        % 重置图片计数器
\renewcommand{\thefigure}{A.\arabic{figure}} % 设置图片编号格式为 A.1, A.2...

\setcounter{equation}{0}      % 重置公式计数器
\renewcommand{\theequation}{A.\arabic{equation}} % 设置公式编号格式为 (A.1), (A.2)...
% --- 附录标号初始化结束 ---

\onecolumn

\section{RLVR Algorithms}
\label{app:algorithms}

In this section, we provide the implementation details of the baseline algorithms discussed in the main text, including Proximal Policy Optimization (PPO)~\cite{schulman2017proximal}, Group Relative Policy Optimization (GRPO)~\cite{shao2024deepseekmath}, REINFORCE~\cite{williams1992simple}, and W-REINFORCE~\cite{zhu2025surprising}.

\subsection{Proximal Policy Optimization (PPO)}

PPO~\cite{schulman2017proximal} constrains the policy update within a trust region by clipping the probability ratio to ensure training stability. Following the notation in Eq.~(\ref{obj:grpo}), the objective function is formulated as:
\begin{equation}
    \mathcal{J}_{\text{PPO}}(\theta) = \mathbb{E}_{x \sim \mathcal{D}} \left[ \frac{1}{G} \sum_{i=1}^G \frac{1}{|\mathbf{y}_i|} \sum_{t=1}^{|\mathbf{y}_i|} \left( \min \left( \rho_{i,t}(\theta) \hat{A}_{i,t}, \text{clip}(\rho_{i,t}(\theta), 1-\epsilon, 1+\epsilon) \hat{A}_{i,t} \right) - \beta \mathbb{D}_{\text{KL}} \right) \right],
    \label{obj:ppo}
\end{equation}
where the term $\mathbb{D}_{\text{KL}}$ (short for $\mathbb{D}_{\text{KL}}(\pi_\theta(y_{i,t}|x, \mathbf{y}_{i,<t}) || \pi_{\text{ref}}(y_{i,t}|x, \mathbf{y}_{i,<t}))$) is the per-token KL divergence penalty used to regularize the policy towards the reference model $\pi_{\text{ref}}$, and $\rho_{i,t}(\theta) = \frac{\pi_\theta(y_{i,t}|x, \mathbf{y}_{i,<t})}{\pi_{\theta_{\text{old}}}(y_{i,t}|x, \mathbf{y}_{i,<t})}$ denotes the probability ratio.

\subsection{Group Relative Policy Optimization (GRPO)}
\label{app:grpo}

GRPO~\cite{shao2024deepseekmath} eliminates the need for a separate critic network by utilizing group-based relative rewards. For each prompt $x$, a group of $G$ responses $\{\mathbf{y}_1, \mathbf{y}_2, \dots, \mathbf{y}_G\}$ is sampled from the old policy $\pi_{\theta_{\text{old}}}$. The objective function is defined as:
\begin{equation}
    \mathcal{J}_{\text{GRPO}}(\theta) = \mathbb{E}_{x \sim \mathcal{D}} \left[ \frac{1}{G} \sum_{i=1}^G \frac{1}{|\mathbf{y}_i|} \sum_{t=1}^{|\mathbf{y}_i|} \left( \min \left( \rho_{i,t}(\theta) \hat{A}_{i,t}, \text{clip}(\rho_{i,t}(\theta), 1-\epsilon, 1+\epsilon) \hat{A}_{i,t} \right) - \beta \mathbb{D}_{\text{KL}} \right) \right].
\end{equation}
The advantage $\hat{A}_{i,t}$ is computed by normalizing the rewards within the group, which reflects the relative quality of the $i$-th response compared to other samples in the same group:
\begin{equation}
    \hat{A}_{i,t} = \frac{r_i - \text{mean}(r_1, \dots, r_G)}{\text{std}(r_1, \dots, r_G) + \epsilon_{\text{std}}},
\end{equation}
where $r_i$ is the outcome-based reward for response $\mathbf{y}_i$, and $\epsilon_{\text{std}}$ is a small constant for numerical stability. Similar to PPO, $\mathbb{D}_{\text{KL}}$ represents the per-token KL divergence between the current policy $\pi_\theta$ and the reference model $\pi_{\text{ref}}$, explicitly added to prevent the policy from deviating too far from the initial distribution.

\subsection{REINFORCE and Weighted-REINFORCE}
\label{app:reinforce}

REINFORCE~\cite{williams1992simple} is a classic policy gradient algorithm that optimizes the model by maximizing the likelihood of positive trajectories and minimizing that of negative ones. Based on the decomposition of the RLVR objective, we define Weighted-REINFORCE (W-REINFORCE)~\cite{zhu2025surprising} by introducing a coefficient $\lambda$ to balance the Positive Sample Reinforcement ($\mathcal{J}_{\text{PSR}}$) and the Negative Sample Reinforcement ($\mathcal{J}_{\text{NSR}}$). Following the group sampling notation, the objective function is formulated as:
\begin{equation}
    \mathcal{J}_{\text{W-REINFORCE}}(\theta) = \mathbb{E}_{x \sim \mathcal{D}} \left[ \frac{1}{G} \sum_{i=1}^G \left( \underbrace{ \mathbb{I}(\mathbf{y}_i \in \mathcal{Y}^+) \cdot \lambda \log \pi_\theta(\mathbf{y}_i|x) }_{\lambda \cdot \mathcal{J}_{\text{PSR}}(\theta)} + \underbrace{ \mathbb{I}(\mathbf{y}_i \in \mathcal{Y}^-) \cdot (-1) \log \pi_\theta(\mathbf{y}_i|x) }_{\mathcal{J}_{\text{NSR}}(\theta)} \right) \right],
\end{equation}
where $\mathcal{Y}^+$ and $\mathcal{Y}^-$ denote the sets of correct and incorrect responses, respectively, and $\mathbb{I}(\cdot)$ is the indicator function. The coefficient $\lambda$ acts as a scaling factor for the positive reinforcement term. This formulation effectively assigns a reward of $+\lambda$ to correct responses and $-1$ to incorrect ones. The standard REINFORCE objective is recovered when $\lambda = 1$. In our experiments, we set $\lambda = 0.1$ to enforce an NSR-dominant update, preventing the reasoning boundary from shrinking while maintaining the error-correction capability.

\subsection{Asymmetric Group Policy Optimization (AGPO)}
\label{app:agpo}

AGPO addresses the reasoning boundary shrinkage problem by introducing an asymmetric advantage estimation mechanism. While it shares the group-based objective framework with GRPO (as shown in Eq.~(\ref{obj:grpo})), it fundamentally redefines how advantages are calculated to decouple positive and negative reinforcement. The objective function is formulated as:
\begin{equation}
    \mathcal{J}_{\text{AGPO}}(\theta) = \mathbb{E}_{x \sim \mathcal{D}} \left[ \frac{1}{G} \sum_{i=1}^G \frac{1}{|\mathbf{y}_i|} \sum_{t=1}^{|\mathbf{y}_i|} \left( \min \left( \rho_{i,t}(\theta) \hat{A}_i^{\text{AGPO}}, \text{clip}(\rho_{i,t}(\theta), 1-\epsilon, 1+\epsilon) \hat{A}_i^{\text{AGPO}} \right) - \beta \mathbb{D}_{\text{KL}} \right) \right],
\end{equation}
where $\rho_{i,t}(\theta)$ is the probability ratio defined in Eq.~(\ref{obj:grpo}), and $\epsilon$ is the clipping parameter. Unlike standard GRPO, the advantage $\hat{A}_i^{\text{AGPO}}$ applies uniformly to the entire sequence (omitting subscript $t$) and is defined as:
\begin{equation}
    \hat{A}_i^{\text{AGPO}} = \underbrace{ \frac{1}{\sqrt{\sigma^2 + \delta^2}} \cdot (r_i - \mu) }_{\text{Constrained group relative term}} + \underbrace{ \mathbb{I}(r_i < 0) \cdot \mathcal{R} }_{\text{Gated negative term}},
\end{equation}
where $r_i \in \{-1, +1\}$ represents the transformed reward for response $\mathbf{y}_i$ (with $-1$ indicating an incorrect outcome), and $\mu$ and $\sigma$ denote the mean and standard deviation of the group rewards, respectively.

\section{Experimental Setup}
\label{app:expset}

\subsection{Experimental Hyperparameters}
\label{app:exp_para}

\textbf{Mathematical Reasoning.}
To ensure a fair comparison, we maintain consistent core configurations across all reinforcement learning algorithms. Our parameter settings generally follow the standard protocols established in recent studies~\cite{yue2025does,zhu2025surprising}. The specific hyperparameters are summarized in Table~\ref{tab:hyperparams}.

\begin{table}[h]
\centering
\caption{Experimental hyperparameters for RLVR baselines and AGPO.}
\label{tab:hyperparams}
\begin{tabular}{lccccc}
\toprule
\textbf{Hyperparameter} & \textbf{PPO} & \textbf{GRPO} & \textbf{REINFORCE} & \textbf{W-REINFORCE} & \textbf{AGPO} \\ \midrule
\texttt{Actor learning rate} & $1 \times 10^{-6}$ & $1 \times 10^{-6}$ & $1 \times 10^{-6}$ & $1 \times 10^{-6}$ & $1 \times 10^{-6}$ \\
\texttt{Critic learning rate} & $1 \times 10^{-5}$ & - & - & - & - \\
\texttt{KL coefficient} ($\beta$) & $0.001$ & $0.005$ & - & - & $0.0$ \\
\texttt{Positive weight} ($\lambda$) & - & - & $1.0$ & $0.1$ & - \\
\texttt{Negative weight} ($\mathcal{R}$) & - & - & $-1.0$ & $-1.0$ & $-1.0$ \\
\texttt{Constraint factor} ($\delta$) & - & - & - & - & $2.0$ \\
\texttt{Train batch size} & $1,024$ & $1,024$ & $1,024$ & $1,024$ & $1,024$ \\
\texttt{Mini-batch size} & $256$ & $256$ & $256$ & $256$ & $256$ \\
\texttt{Rollout group size} ($G$) & $8$ & $8$ & $8$ & $8$ & $8$ \\
\texttt{Max prompt length} & $1,024$ & $1,024$ & $1,024$ & $1,024$ & $1,024$ \\
% Max Response Length & $3072$ & $3072$ & $3072$ & $3072$ & $3072$ \\
\texttt{Total epochs} & $20$ & $20$ & $20$ & $20$ & $20$ \\ \bottomrule
\end{tabular}
\end{table}

The weight factor $\lambda$ is used to adjust the contribution of the positive reinforcement term in REINFORCE variants. For AGPO, we introduce the constraint factor $\delta$ and the negative reward weight $\mathcal{R}$ to control the asymmetric advantage estimation. For all algorithms, we set the number of rollouts per prompt ($G$) to 8 to ensure consistent computational budget and data exposure. We use a temperature of 0.6 for all models. Moreover, regarding the generation configuration, the maximum response length is set based on the model's capability. Therefore, we use a length of 3,072 for Qwen2.5-Math-7B~\cite{yang2024qwen2} and Llama-3.1-8B-Instruct~\cite{grattafiori2024llama}, whereas for Qwen3-4B~\cite{yang2025qwen3}, we extend this limit to 31,744 to accommodate its longer context window.

\textbf{Search Ads Relevance Reasoning.}

\begin{figure}[h!]
    \centering
    \includegraphics[width=0.52\linewidth]{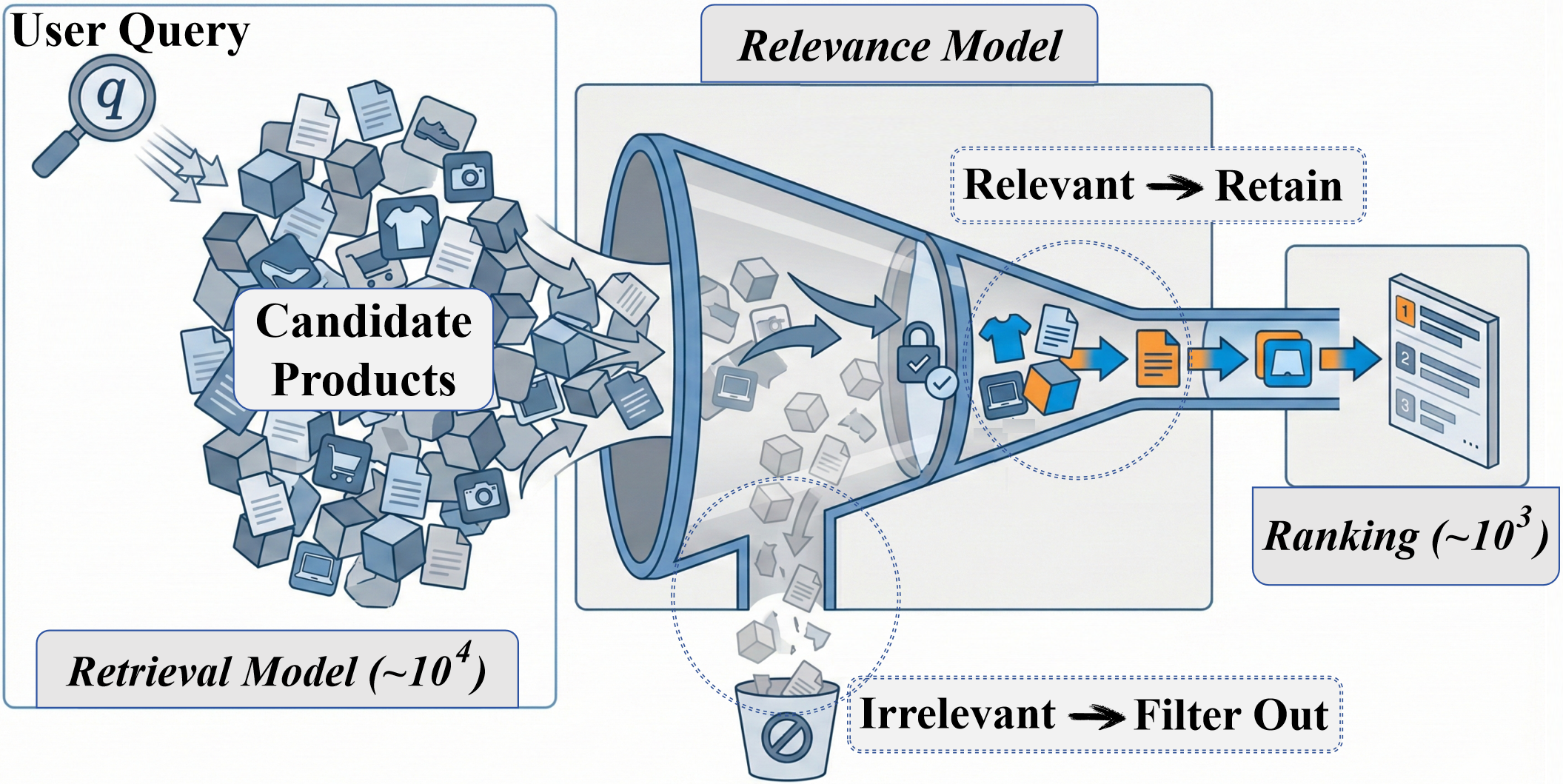}
    \caption{Overview of the relevance model in the search ads funnel, filtering out mismatched candidates before the ranking stage.}
    \label{fig:relevance_model}
\end{figure}

In large-scale \textit{e-commerce search advertising systems}~\cite{fan2019mobius,liu2021que2search}, the relevance model serves as the critical \text{gatekeeper} that filters out semantically mismatched advertisements before entering the ranking stage~\cite{lu2020twinbert}.
As illustrated in Figure~\ref{fig:relevance_model}, this module operates downstream of candidate retrieval, ensuring that only semantically aligned advertisements proceed to the ranking stage.
It should be noted that only the relevance model (a BERT model) was replaced in the online A/B test, while the other modules in the advertising system remained unchanged.

In industrial application experiments, we use Rele-Ads-8B, a decoder-only transformer model designed for semantic relevance judgment in JD Ads Search. This model functions as a teacher to generate high-quality annotation data enriched with explicit CoT reasoning processes.
To ensure a rigorous evaluation of the reasoning capability boundary, we maintain a consistent generation configuration across all experiments. Specifically, for both the base Rele-Ads-8B model and its optimized variants (trained by AGPO and GRPO), we employ a sampling strategy with a temperature of $0.6$, a top-$p$ of $0.9$, and a top-$k$ of $40$. The maximum generation length is set to 4096, which provides a sufficient context window for the model to perform a step-by-step semantic analysis and output the final relevance label. The student BERT model (3 layers, 256 hidden units) was trained on 40 million online query-product pairs. For the experimental evaluation, a new BERT model ($\text{BERT}_\textbf{+AGPO Labels}$) was trained on the same dataset using relevance labels re-annotated by the AGPO-optimized Rele-Ads-8B teacher model. In the production system, only query-product pairs predicted as ``fully relevant" by the student BERT model could be presented to the users.

\subsection{Evaluation Metrics}
\label{app:eva}

\textbf{Pass@\textit{k}.}
We employ the unbiased estimator proposed by Chen et al.~\cite{chen2021evaluating} to evaluate the reasoning capability boundary. Given $n$ generated samples per problem and $c$ correct samples among them, the unbiased Pass@$k$ is calculated as:
\begin{equation}
    \text{Pass}@k = \mathbb{E} \left[ 1 - \frac{\binom{n-c}{k}}{\binom{n}{k}} \right].
\end{equation}
We adopt a full spectrum of Pass@$k$ as our main evaluation metric, using $k\in\{1,2,4,8,16,32,64,128,256\}$ for Qwen2.5-Math-7B, Qwen3-4B, and Llama-3.1-8B-Instruct.

\textbf{Proportion of Irrelevant Products (PIR).}
For the Search Ads Relevance task, PIR serves as a critical metric for evaluating the model's filtering effectiveness. In the context of Search Ads, the model categorizes each query-product pair into a relevance level $y \in \{0, 1, 2\}$, representing \textit{irrelevant}, \textit{partially relevant}, and \textit{fully relevant}, respectively. PIR quantifies the frequency of ``bad cases" where the model over-predicts relevance. Formally, let $\mathcal{D}$ be the evaluation dataset and $\hat{y}_i$ be the predicted label for instance $i$. We define the set of fully relevant predictions as $\mathcal{S}^{+} = \{i \in \mathcal{D} \mid \hat{y}_i = 2\}$. The PIR is then defined as:
\begin{equation}
    \text{PIR} = \frac{\sum_{i \in \mathcal{S}^{+}} \mathbb{I}(y_i = 0)}{|\mathcal{S}^{+}|},
\end{equation}
where $y_i$ is the ground-truth label assigned by human annotators and $\mathbb{I}(\cdot)$ is the indicator function. 
In the advertising system, ads that are classified as fully relevant by the model will eventually be shown to the user.
A lower PIR indicates superior semantic alignment, as it reflects the model's ability to minimize the exposure of irrelevant advertisements that were erroneously classified as fully relevant.

\textbf{Click-Through Rate Per Impression (CTRPI).} This metric measures the probability that an advertisement is clicked given its impression. For a set of events $\mathcal{E}$, CTRPI is defined as~\cite{richardson2007predicting}
\begin{equation}
    \text{CTR}_{\text{PI}} = \frac{\sum_{e \in \mathcal{E}} \mathbb{I}(\text{click}_e)}{\sum_{e \in \mathcal{E}} \mathbb{I}(\text{impression}_e)}.
\end{equation}
It serves as a direct proxy for the semantic relevance between the query and the advertisement from the user's perspective.

\textbf{Cost Per Mille (CPM).} This metric, also referred to as effective Cost Per Mille (eCPM) in display advertising, represents the expected revenue generated per 1,000 impressions. To formally define CPM, we first denote Cost Per Click (CPC) as the average revenue accumulated per click
\begin{equation}
    \text{CPC} = \frac{\sum_{e \in \mathcal{E}} \text{revenue}_e}{\sum_{e \in \mathcal{E}} \mathbb{I}(\text{click}_e)}.
\end{equation}
where $\text{revenue}_e$ is the cost paid by the advertiser for event $e$. CPM is then derived from the relationship between click-through rate and CPC~\cite{zhu2017optimized}
\begin{equation}
    \text{CPM} = 1000 \times \text{CTR}_{\text{PI}} \times \text{CPC}.
\end{equation}
This identity captures the platform's monetization efficiency, as it evaluates the commercial value produced by the allocated traffic across every thousand impressions.

\textbf{Gross Merchandise Volume (GMV).} In the operational standards in e-commerce search, GMV quantifies the total economic value of transactions. For a period $T$, let $P_i$ and $Q_i$ be the price and quantity of the $i$-th purchased item respectively; GMV is calculated as~\cite{liu2017cascade}
\begin{equation}
    \text{GMV} = \sum_{i \in \text{Purchases}_T} P_i \times Q_i.
\end{equation}
It indicates the overall transaction scale and the health of the downstream e-commerce ecosystem.

\subsection{Prompt Templates}
\label{app:prompt}

\textbf{Mathematical Reasoning.}
We utilize specific prompt templates tailored to the instruction-tuning formats of different models, following standard practices~\cite{zeng2025simplerl,yue2025does,zhu2025surprising}. The templates for Qwen2.5-Math-7B, Qwen3-4B, and Llama-3.1-8B-Instruct are presented in Tables~\ref{tab:prompt_qwen25}, \ref{tab:prompt_qwen3}, and \ref{tab:prompt_llama}, respectively.

% --- Table 1: Qwen2.5-Math-7B ---
\begin{table}[h!]
    \centering
    \caption{Prompt template for Qwen2.5-Math-7B.}
    \label{tab:prompt_qwen25}
    \begin{tcolorbox}[
        colback=black!5!white,
        colframe=black,
        boxrule=0.8pt,
        arc=3mm,
        width=0.95\textwidth,
        left=6pt, right=6pt, top=6pt, bottom=6pt
    ]
        \setlength{\parindent}{0pt}
        \ttfamily % 使用打字机字体，看起来更像代码/Prompt
        \small
        \textless\textbar im\_start\textbar\textgreater system \\
        You are a helpful assistant.\textless\textbar im\_end\textbar\textgreater \\
        \textless\textbar im\_start\textbar\textgreater user \\
        \{input\} \\
        Please reason step by step, and put your final answer within \textbackslash boxed\{\}.\textless\textbar im\_end\textbar\textgreater \\
        \textless\textbar im\_start\textbar\textgreater assistant
    \end{tcolorbox}
\end{table}

% --- Table 2: Qwen3-4B (No Thinking) ---
\begin{table}[h!]
    \centering
    \caption{Prompt template for Qwen3-4B (Non-thinking Mode).}
    \label{tab:prompt_qwen3}
    \begin{tcolorbox}[
        colback=black!5!white,
        colframe=black,
        boxrule=0.8pt,
        arc=3mm,
        width=0.95\textwidth,
        left=6pt, right=6pt, top=6pt, bottom=6pt
    ]
        \setlength{\parindent}{0pt}
        \ttfamily
        \small
        \textless\textbar im\_start\textbar\textgreater system \\
        You are a helpful assistant.\textless\textbar im\_end\textbar\textgreater \\
        \textless\textbar im\_start\textbar\textgreater user \\
        \{input\} \\
        Please reason step by step, and put your final answer within \textbackslash boxed\{\}.\textless\textbar im\_end\textbar\textgreater \\
        \textless\textbar im\_start\textbar\textgreater assistant \\
        \textless think\textgreater \\
        \\
        \textless/think\textgreater
    \end{tcolorbox}
\end{table}

% --- Table 3: Llama-3.1-8B-Instruct ---
\begin{table}[h!]
    \centering
    \caption{Prompt template for Llama-3.1-8B-Instruct.}
    \label{tab:prompt_llama}
    \begin{tcolorbox}[
        colback=black!5!white,
        colframe=black,
        boxrule=0.8pt,
        arc=3mm,
        width=0.95\textwidth,
        left=6pt, right=6pt, top=6pt, bottom=6pt
    ]
        \setlength{\parindent}{0pt}
        \ttfamily
        \small
        \textless\textbar begin\_of\_text\textbar\textgreater\textless\textbar start\_header\_id\textbar\textgreater system\textless\textbar end\_header\_id\textbar\textgreater \\
        \\
        You are a helpful assistant.\textless\textbar eot\_id\textbar\textgreater\textless\textbar start\_header\_id\textbar\textgreater user\textless\textbar end\_header\_id\textbar\textgreater \\
        \\
        \{input\} \\
        Please reason step by step, and put your final answer within \textbackslash boxed\{\}.\textless\textbar eot\_id\textbar\textgreater\textless\textbar start\_header\_id\textbar\textgreater assistant\textless\textbar end\_header\_id\textbar\textgreater
    \end{tcolorbox}
\end{table}

\textbf{Search Ads Relevance Reasoning.}
Given that our industrial application is deployed on JD.com, a leading Chinese e-commerce platform, the prompts used in our experiments are originally in Chinese. To ensure reproducibility and accessibility for a broader audience, we present the original Chinese prompts followed by their corresponding English translations in Table~\ref{tab:prompt_ads}. The instruction string combines the role definition, specific instance data (query and sku), and strict output formatting constraints.

\begin{table}[h!]
    \centering
    \caption{Prompt template for Search Ads Relevance Reasoning (Chinese Original with English Translation).}
    \label{tab:prompt_ads}
    \begin{tcolorbox}[
        colback=black!5!white,
        colframe=black,
        boxrule=0.8pt,
        arc=3mm,
        width=0.95\textwidth,
        left=6pt, right=6pt, top=6pt, bottom=6pt
    ]
        \setlength{\parindent}{0pt}
        \ttfamily
        \small
        % --- 开启中文环境 ---
        \begin{CJK*}{UTF8}{gbsn}
        
        % --- Original Chinese Section ---
        \textbf{[Original Chinese Prompt]} \\
        \textless\textbar im\_start\textbar\textgreater user \\
        你是一个query-sku相关性标注人员, 需要判别给定的query和sku相关性类型. \\
        当query是"\{query\}", sku是"\{sku\_name\}", \\
        综合产品词匹配情况和关键属性匹配情况, 判断query和sku属于\textless不相关\textgreater、\textless完全相关\textgreater、\textless部分相关\textgreater哪一类。\\
        注意一定要以属于-{}-\textgreater对应类型结束，并输出分析: \\
        \textless\textbar im\_end\textbar\textgreater \\
        \textless\textbar im\_start\textbar\textgreater assistant \\
        
        \end{CJK*}
        % --- 关闭中文环境 ---
        
        \vspace{1em}
        \hrule
        \vspace{1em}
        
        % --- English Translation Section ---
        \textbf{[English Translation]} \\
        \textless\textbar im\_start\textbar\textgreater user \\
        You are a query-sku relevance annotator, required to determine the relevance type between a given query and sku. \\
        When the query is "\{query\}", and the sku is "\{sku\_name\}", \\
        comprehensively consider the product term matching and key attribute matching to judge which category the query and sku belong to: \textless Irrelevant\textgreater, \textless Fully Relevant\textgreater, or \textless Partially Relevant\textgreater. \\
        Note that you must end with "belongs to -{}-\textgreater corresponding type", and output the analysis: \\
        \textless\textbar im\_end\textbar\textgreater \\
        \textless\textbar im\_start\textbar\textgreater assistant
    \end{tcolorbox}
\end{table}

\newpage

\section{More Results on Mathematics}
\label{app:morere}

\begin{figure*}[t]
    \centering
    \includegraphics[width=0.68\textwidth]{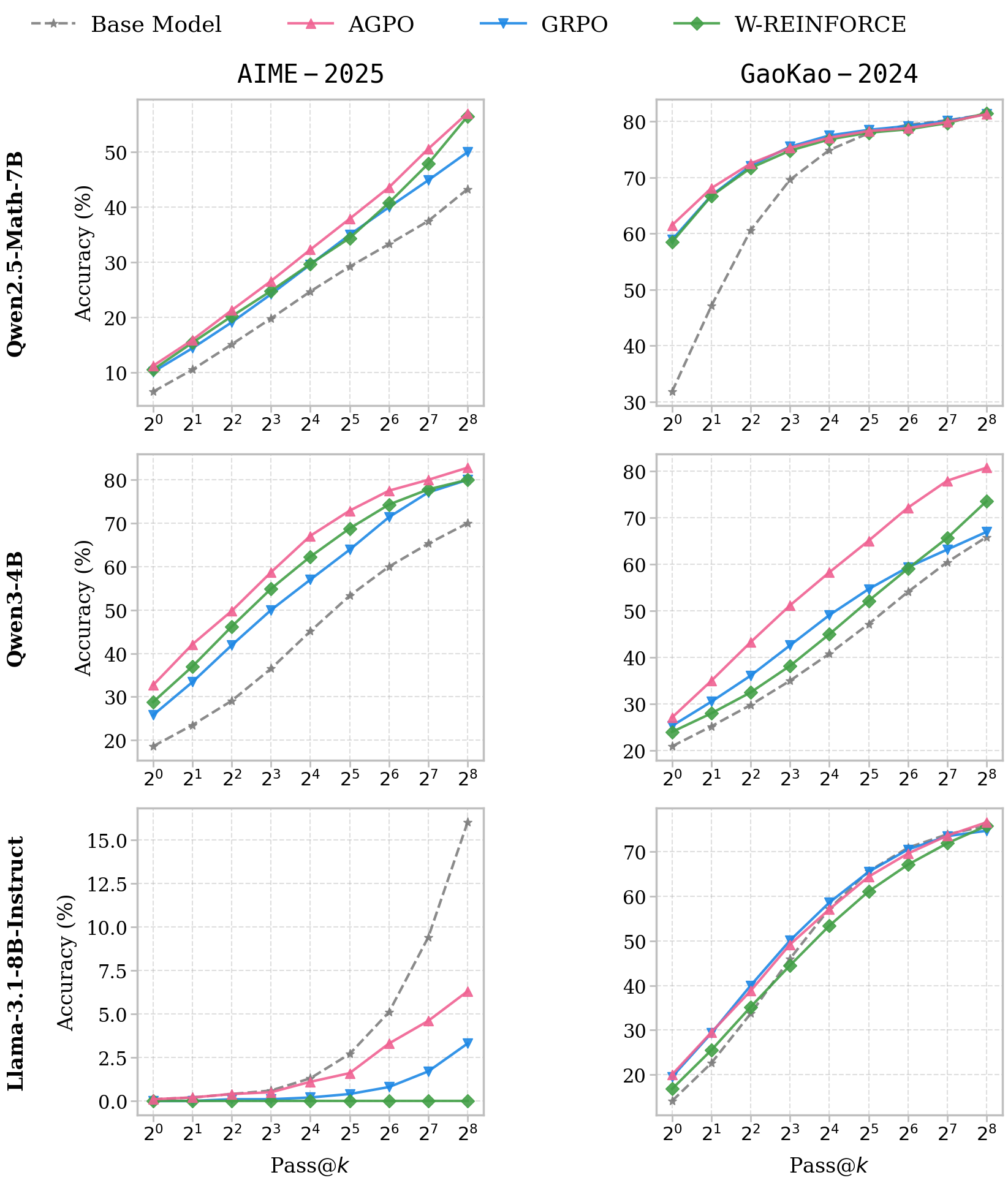}
    \caption{Pass@$k$ performance scaling on \texttt{AIME-2025} and \texttt{GaoKao-2024} benchmarks with Qwen2.5-Math-7B, Qwen3-4B and Llama-3.1-8B-Instruct models.
    }
    \label{fig:appendix_passk}
\end{figure*}

We extend our evaluation to include two additional benchmarks to further evaluate the generalization of AGPO: \texttt{AIME-2025} and \texttt{GaoKao-2024}. These results, conducted across Qwen2.5-Math-7B, Qwen3-4B, and Llama-3.1-8B-Instruct models, are shown in Figure~\ref{fig:appendix_passk}. Overall, AGPO consistently maintains its lead over other RLVR baselines, demonstrating superior scaling trends.
On the \texttt{GaoKao-2024} benchmark, AGPO exhibits clear advantages across all tested models. Particularly with the Qwen3-4B model, AGPO achieves a substantial performance lead over GRPO and W-REINFORCE across the entire Pass@$k$ spectrum. This indicates that AGPO's asymmetric update strategy is highly effective at mining the model's latent reasoning potential while successfully preserving the reasoning boundary.
In particular, a notable performance bottleneck is observed for all RL methods when applying Llama-3.1-8B-Instruct to the \texttt{AIME-2025} dataset. As illustrated in the bottom-left panel of Figure~\ref{fig:appendix_passk}, the initial Pass@$1$ accuracy of the base model on this task is virtually zero. This leads to a severe ``cold-start'' problem in the RLVR training paradigm. Since algorithms like AGPO and GRPO rely on the identification and amplification of correct reasoning paths found within the on-policy sampling distribution, the near-total absence of successful trajectories during the initial stages of training prevents the model from accumulating sufficient positive reward signals. Consequently, the policy is unable to establish effective reasoning patterns, resulting in stagnant performance that remains significantly below the base model's latent coverage boundary (Pass@$256$). This observation underscores that the effectiveness of current RLVR methods is inherently bounded by the base model's initial capacity to discover correct solutions within the sampling budget.

\newpage

\section{Impact of Parameters in AGPO}
\label{app:hyperparameter_analysis}

In this section, we evaluate the hyperparameters setting in AGPO, \textit{i.e.}, the constraint factor $\delta$ and the KL coefficient $\beta$. All experiments are conducted using the Qwen2.5-Math-7B model.

\subsection{Impact of the Constraint Factor $\delta$}
The constraint factor $\delta$ regulates the magnitude of the group-relative advantage in AGPO. To isolate its effect, we fix $\beta=0$ and test $\delta \in \{0.0001, 0.5, 1, 2, 5\}$. The results are shown in Table~\ref{tab:delta_ablation}.
As $\delta$ increases, the strategy becomes increasingly \textit{NSR-dominated}. 
When $\delta$ approaches infinity, AGPO degenerates into a pure NSR~\cite{zhu2025surprising}.
Conversely, as $\delta$ approaches $0$, the advantage estimation approaches standard GRPO while retaining a stable negative bias. Overall, $\delta=2$ is the optimal parameter setting in practice.

\begin{table}[htbp]
\centering
% \caption{Impact of $\delta$ on \texttt{Qwen2.5-Math-7B} ($\beta=0$). Results are reported in (\%).}
\caption{Pass@$k$ (\%) results on Qwen2.5-Math-7B with different $\delta$ values.}
\label{tab:delta_ablation}
% \footnotesize
\setlength{\tabcolsep}{2pt}
\begin{tabular}{l c@{\hspace{6pt}}c c@{\hspace{6pt}}c c@{\hspace{6pt}}c c@{\hspace{6pt}}c c@{\hspace{6pt}}c}
\toprule
& \multicolumn{2}{c}{\texttt{MATH}} & \multicolumn{2}{c}{\texttt{Olympiad}} & \multicolumn{2}{c}{\texttt{AIME-2024}} & \multicolumn{2}{c}{\texttt{AIME-2025}} & \multicolumn{2}{c}{\texttt{GaoKao-2024}} \\
\cmidrule(lr){2-3} \cmidrule(lr){4-5} \cmidrule(lr){6-7} \cmidrule(lr){8-9} \cmidrule(lr){10-11}
\textbf{Setting} & Pass@$1$ & Pass@$256$ & Pass@$1$ & Pass@$256$ & Pass@$1$ & Pass@$256$ & Pass@$1$ & Pass@$256$ & Pass@$1$ & Pass@$256$ \\
\midrule
$\delta = 0.0001$ & 75.7 & 95.1 & \underline{38.3} & 72.3 & 27.6 & 73.6 & \underline{10.7} & 51.2 & 58.6 & \underline{80.6} \\
$\delta = 0.5$    & 75.6 & 94.9 & 37.8 & 73.5 & 28.8 & 73.3 & 9.9 & 53.7 & 58.9 & 80.0 \\
$\delta = 1.0$    & \underline{76.8} & 95.7 & 38.2 & 73.4 & \underline{29.5} & 75.5 & 10.3 & 55.8 & \textbf{61.7} & 80.0 \\
\textbf{$\delta = 2.0$} & \textbf{77.2} & \underline{96.6} & \textbf{38.6} & \textbf{75.3} & \textbf{30.1} & \textbf{76.8} & \textbf{11.2} & \underline{57.1} & \underline{61.5} & \textbf{81.3} \\
$\delta = 5.0$    & 76.5 & \textbf{96.7} & 36.4 & \underline{75.0} & 28.3 & \underline{76.4} & 10.1 & \textbf{57.4} & 59.0 & \textbf{81.3} \\
\bottomrule
\end{tabular}
\end{table}

\subsection{Impact of the KL Coefficient $\beta$}
We investigate the influence of the KL divergence penalty by fixing $\delta=2$ and varying $\beta \in \{0, 0.0005, 0.001\}$. As shown in Table~\ref{tab:beta_ablation}, AGPO is robust to removal of the KL penalty. 
Consistent with the analysis in  W-REINFORCE~\cite{zhu2025surprising}, when the policy is dominated by NSR, the policy no longer depends on KL constraints as it does with GRPO and PPO.
It is worth noting that when $\beta=0$, we observed a similar situation during training to that of NSR~\cite{zhu2025surprising}, where extensive training (\textit{e.g.}, over hundreds of gradient steps) leads to a noticeable decline in performance. However, setting $\beta=0$ remains the optimal parameter setting in practice.

\begin{table}[htbp]
\centering
\caption{Pass@$k$ (\%) results on Qwen2.5-Math-7B with different $\beta$ values.}
\label{tab:beta_ablation}
% \footnotesize
\setlength{\tabcolsep}{2pt}
\begin{tabular}{l c@{\hspace{6pt}}c c@{\hspace{6pt}}c c@{\hspace{6pt}}c c@{\hspace{6pt}}c c@{\hspace{6pt}}c}
\toprule
& \multicolumn{2}{c}{\texttt{MATH}} & \multicolumn{2}{c}{\texttt{Olympiad}} & \multicolumn{2}{c}{\texttt{AIME-2024}} & \multicolumn{2}{c}{\texttt{AIME-2025}} & \multicolumn{2}{c}{\texttt{GaoKao-2024}} \\
\cmidrule(lr){2-3} \cmidrule(lr){4-5} \cmidrule(lr){6-7} \cmidrule(lr){8-9} \cmidrule(lr){10-11}
\textbf{Setting} & Pass@$1$ & Pass@$256$ & Pass@$1$ & Pass@$256$ & Pass@$1$ & Pass@$256$ & Pass@$1$ & Pass@$256$ & Pass@$1$ & Pass@$256$ \\
\midrule
\textbf{$\beta = 0$} & \textbf{77.2} & \textbf{96.6} & \textbf{38.6} & \textbf{75.3} & \textbf{30.1} & \textbf{76.8} & \textbf{11.2} & \textbf{57.1} & \textbf{61.5} & \textbf{81.3} \\
$\beta = 0.0005$ & \underline{76.8} & 96.0 & \underline{38.0} & \underline{74.5} & \underline{29.2} & 76.4 & \underline{10.3} & \textbf{57.1} & \underline{60.8} & \textbf{81.3} \\
$\beta = 0.001$  & 76.3 & \underline{96.3} & 37.4 & 74.4 & 28.5 & \underline{76.6} & 10.0 & \underline{56.6} & 60.1 & \textbf{81.3} \\
\bottomrule
\end{tabular}
\end{table}

\newpage

\section{CoT Case Analysis}

\subsection{Mathematical Reasoning Cases}

\begin{figure}[h!]
\centering
% --- Question Box ---
\begin{tcolorbox}[
    colback=yellow!5!white, 
    colframe=yellow!50!black, 
    title=\textbf{Question 1 (Combinatorial Partitioning)},
    left=6pt, right=6pt, top=6pt, bottom=6pt, 
    fonttitle=\bfseries
]
    Determine the greatest positive integer $k$ that satisfies the following property: The set of positive integers can be partitioned into $k$ subsets $A_{1}, A_{2}, \ldots, A_{k}$ such that for all integers $n \geq 15$ and all $i \in \{1,2, \ldots, k\}$ there exist two distinct elements of $A_{i}$ whose sum is $n$.
\end{tcolorbox}

% --- Full AGPO CoT Box ---
\begin{tcolorbox}[
    colback=purple!5!white, 
    colframe=purple!50!black, 
    title=\textbf{AGPO-optimized Qwen2.5-Math-7B Correct Reasoning Path \hfill },
    left=6pt, right=6pt, top=6pt, bottom=6pt, 
    fonttitle=\bfseries
]
    To solve this problem, we need to find the greatest positive integer $k$ such that the set of positive integers can be partitioned into $k$ subsets $A_1, A_2, \ldots, A_k$ with the property that for all integers $n \geq 15$ and all $i \in \{1, 2, \ldots, k\}$, there exist two distinct elements of $A_i$ whose sum is $n$.

    Let's analyze the problem step by step:

    \textbf{1. Understanding the Problem:} For each subset $A_i$, we need to ensure that for any $n \geq 15$, there exist two distinct elements $a, b \in A_i$ such that $a + b = n$. This means that each subset $A_i$ must contain pairs of numbers that sum up to all integers starting from 15.

    \textbf{2. Constructing Subsets:} To achieve this, we need to carefully construct the subsets. One way to do this is to consider the parity (odd or even) of the numbers. For example, we can partition the positive integers into two subsets:
    \begin{itemize}
        \item $A_1$ contains all odd numbers.
        \item $A_2$ contains all even numbers.
    \end{itemize}
    However, this simple partition doesn't work because for $n = 15$, we need two distinct elements from the same subset to sum to 15. For $A_1$, the possible pairs are (1, 14), (3, 12), (5, 10), (7, 8), but 14, 12, 10, and 8 are not in $A_1$. Similarly, for $A_2$, the possible pairs are (2, 13), (4, 11), (6, 9), (8, 7), but 13, 11, and 9 are not in $A_2$.

    \textbf{3. Further Partitioning:} We need to further partition the subsets to ensure that for each $n \geq 15$, there exist two distinct elements in the same subset that sum to $n$. One possible way is to partition the positive integers into three subsets:
    \begin{itemize}
        \item $A_1$ contains numbers of the form $3k+1$.
        \item $A_2$ contains numbers of the form $3k+2$.
        \item $A_3$ contains numbers of the form $3k$.
    \end{itemize}
    Let's check if this works: For $n = 15$, we can take (1, 14), (4, 11), (7, 8) from $A_1$, (2, 13), (5, 10), (8, 7) from $A_2$, and (3, 12), (6, 9), (9, 6) from $A_3$. Notice that for any $n \geq 15$, we can always find pairs in the same subset that sum to $n$.

    \textbf{4. Generalizing:} We can generalize this idea to show that we can partition the positive integers into $k = 3$ subsets and satisfy the condition. If we try to partition into more than 3 subsets, we would need to ensure that for each $n \geq 15$, there exist two distinct elements in the same subset that sum to $n$, which becomes increasingly complex and may not be possible with more than 3 subsets.

    Therefore, the greatest positive integer $k$ that satisfies the condition is $\boxed{3}$.
\end{tcolorbox}
% \caption{Case 1: Qwen2.5-Math-7B (after AGPO training) Correct Response for Combinatorial Partitioning Problem in \texttt{Olympiad}.}
\caption{Case 1: Full correct reasoning trajectory of AGPO-optimized Qwen2.5-Math-7B on a combinatorial partitioning problem from the \texttt{Olympiad} benchmark.}
\label{fig:math_full_cot_7}
\end{figure}

\begin{figure}[htbp]
\centering
% --- Question Box (与 Case 1 保持一致) ---
\begin{tcolorbox}[
    colback=yellow!5!white, 
    colframe=yellow!50!black, 
    title=\textbf{Question 1 (Combinatorial Partitioning)},
    left=6pt, right=6pt, top=6pt, bottom=6pt, 
    fonttitle=\bfseries
]
    Determine the greatest positive integer $k$ that satisfies the following property: The set of positive integers can be partitioned into $k$ subsets $A_{1}, A_{2}, \ldots, A_{k}$ such that for all integers $n \geq 15$ and all $i \in \{1,2, \ldots, k\}$ there exist two distinct elements of $A_{i}$ whose sum is $n$.
\end{tcolorbox}

% --- Full Base Model Incorrect Response Box ---
\begin{tcolorbox}[
    colback=gray!5!white, 
    colframe=gray!50!black, 
    title=\textbf{Qwen2.5-Math-7B Incorrect Reasoning Path \hfill},
    left=6pt, right=6pt, top=6pt, bottom=6pt, 
    fonttitle=\bfseries
]
    To determine the greatest positive integer $k$ that satisfies the given property, we need to partition the set of positive integers into $k$ subsets $A_1, A_2, \ldots, A_k$ such that for all integers $n \geq 15$ and all $i \in \{1, 2, \ldots, k\}$, there exist two distinct elements of $A_i$ whose sum is $n$. 

    Let's break down the problem step-by-step:

    \textbf{1. Understand the Problem:}
    We need to partition the set of positive integers into $k$ subsets, such that each subset $A_i$ contains pairs of distinct elements whose sum is any integer $n \geq 15$.

    \textbf{2. Consider Small Values of $k$:}
    \begin{itemize}
        \item If $k = 1$, then we need to partition all positive integers into one subset $A_1$. Clearly, this is possible since we can always find two distinct elements of $A_1$ whose sum is $n$ for $n \geq 15$.
        \item If $k = 2$, we can partition the positive integers into odd and even numbers. This works since for any $n \geq 15$, we can find two odd numbers or two even numbers whose sum is $n$.
        \item If $k = 3$, we can partition the positive integers into numbers that are $0 \pmod 3$, $1 \pmod 3$, and $2 \pmod 3$. This also works since for any $n \geq 15$, we can find two numbers in the same residue class whose sum is $n$.
        \item If $k = 4$, we can partition the positive integers into numbers that are $0, 1, 2,$ or $3 \pmod 4$. This works similarly.
    \end{itemize}

    \textbf{3. Generalize for $k$:}
    We can generalize this to $k$ subsets, where each subset contains numbers that are congruent to the same residue modulo $k$. For $k \geq 5$, we need to check if it's possible to ensure that for all $n \geq 15$, there exist two distinct elements in each subset whose sum is $n$. However, for $k = 5$, we can't ensure that for all $n \geq 15$, there exist two distinct elements in each subset whose sum is $n$. For example, if we partition the integers into $0, 1, 2, 3,$ and $4 \pmod 5$, then for $n = 15$, the only way to get pairs whose sum is 15 is to use elements from different subsets (like 5 and 10).

    Thus, the greatest positive integer $k$ that satisfies the property is 4. We can confirm this by checking that for $k = 4$, we can partition the integers into $0, 1, 2,$ and $3 \pmod 4$ and ensure that for all $n \geq 15$, there exist two distinct elements in each subset whose sum is $n$.

    Let's write the final answer in the boxed format: $\boxed{4}$.
\end{tcolorbox}
\caption{Case 1: Incorrect reasoning trajectory of Qwen2.5-Math-7B on a combinatorial partitioning problem from the \texttt{Olympiad} benchmark.}
\label{fig:math_base_cot_failure}
\end{figure}

\begin{figure}[htbp]
\centering
% --- Question Box ---
\begin{tcolorbox}[
    colback=yellow!5!white, 
    colframe=yellow!50!black, 
    title=\textbf{Question 2 (Geometry and Polynomials)},
    left=6pt, right=6pt, top=6pt, bottom=6pt, 
    fonttitle=\bfseries
]
    Let $n \geq 2$ be an integer, and let $f$ be a $4n$-variable polynomial with real coefficients. Assume that, for any $2n$ points $(x_{1}, y_{1}), \ldots,(x_{2n}, y_{2n})$ in the plane, $f(x_{1}, y_{1}, \ldots, x_{2n}, y_{2n})=0$ if and only if the points form the vertices of a regular $2n$-gon in some order, or are all equal. Determine the smallest possible degree of $f$.
\end{tcolorbox}

% --- Full AGPO CoT Box ---
\begin{tcolorbox}[
    colback=purple!5!white, 
    colframe=purple!50!black, 
    title=\textbf{AGPO-optimized Qwen2.5-Math-7B Correct Reasoning Path \hfill },
    left=6pt, right=6pt, top=6pt, bottom=6pt, 
    fonttitle=\bfseries
]
    To solve this problem, we need to determine the smallest possible degree of the polynomial $f$ that satisfies the given conditions. Let's break down the problem step by step:

    \textbf{1. Understanding the Polynomial}:
    \begin{itemize}
        \item The polynomial $f$ is a $4n$-variable polynomial, meaning it takes $2n$ $x$-coordinates and $2n$ $y$-coordinates as inputs.
        \item $f$ evaluates to zero if and only if the points form the vertices of a regular $2n$-gon in some order or are all equal.
    \end{itemize}

    \textbf{2. Degree of the Polynomial}:
    \begin{itemize}
        \item The polynomial $f$ must be symmetric under rotations and reflections of the regular $2n$-gon.
        \item The polynomial must be able to distinguish between the vertices of a regular $2n$-gon and other configurations of points.
    \end{itemize}

    \textbf{3. Smallest Possible Degree}:
    \begin{itemize}
        \item The smallest degree of $f$ can be determined by considering the properties of a regular $2n$-gon. The polynomial must be able to capture the geometric constraints of the regular $2n$-gon.
        \item The degree of the polynomial is influenced by the fact that it must be able to distinguish between the vertices of a regular $2n$-gon and other configurations. The smallest possible degree that can capture this property is $2n$.
    \end{itemize}

    The polynomial must be of degree at least $2n$ because it needs to encode the geometric constraints of the regular $2n$-gon, which involves $2n$ points. Therefore, the smallest possible degree of $f$ is $2n$.

    Let's put this into code to verify the reasoning:
    \begin{tcolorbox}[colback=white, boxrule=0.5pt, arc=1mm]
    \ttfamily
    import sympy as sp \\
    n = sp.symbols('n', integer=True) \\
    smallest\_degree = 2 * n \\
    print(smallest\_degree)
    \end{tcolorbox}
    \textbf{Output:} \texttt{2*n}

    The smallest possible degree of the polynomial $f$ is $\boxed{2n}$.
\end{tcolorbox}
\caption{Case 2: Full correct reasoning trajectory of AGPO-optimized Qwen2.5-Math-7B on a geometry and polynomials problem from the \texttt{Olympiad} benchmark.}
\label{fig:math_full_cot_case3}
\end{figure}

\begin{figure}[htbp]
\centering
% --- Question Box (Reference Only) ---
\begin{tcolorbox}[
    colback=yellow!5!white, 
    colframe=yellow!50!black, 
    title=\textbf{Question 2 (Geometry and Polynomials)},
    left=6pt, right=6pt, top=6pt, bottom=6pt, 
    fonttitle=\bfseries
]
    Let $n \geq 2$ be an integer, and let $f$ be a $4n$-variable polynomial with real coefficients. Assume that, for any $2n$ points $(x_{1}, y_{1}), \ldots,(x_{2n}, y_{2n})$ in the plane, $f(x_{1}, y_{1}, \ldots, x_{2n}, y_{2n})=0$ if and only if the points form the vertices of a regular $2n$-gon in some order, or are all equal. Determine the smallest possible degree of $f$.
\end{tcolorbox}

% --- Full Base Model Incorrect Response Box ---
\begin{tcolorbox}[
    colback=gray!5!white, 
    colframe=gray!50!black, 
    title=\textbf{Qwen2.5-Math-7B Incorrect Reasoning Path \hfill},
    left=6pt, right=6pt, top=6pt, bottom=6pt, 
    fonttitle=\bfseries
]
    To solve this problem, let's analyze the given conditions. We need to find the smallest possible degree of the polynomial $f$ such that $f(x_1, y_1, \ldots, x_{2n}, y_{2n}) = 0$ if and only if the points $(x_1, y_1), \ldots, (x_{2n}, y_{2n})$ form the vertices of a regular $2n$-gon in some order or are all equal.
    \\

    First, consider the geometric properties of a regular $2n$-gon. The coordinates of the vertices of a regular $2n$-gon centered at the origin with radius $r$ can be written as:
    \[ \left( r \cos\left(\frac{2k\pi}{2n}\right), r \sin\left(\frac{2k\pi}{2n}\right) \right), \quad k = 0, 1, 2, \ldots, 2n-1. \]
    This means the polynomial $f$ must be zero if the given points are permutations of these coordinates or all equal. 
    \\

    Since the polynomial must be zero for all permutations of these coordinates, the polynomial must be symmetric in the $x_i$ and $y_i$ coordinates. The simplest polynomial that is zero for all permutations of these coordinates is the polynomial that is zero when the points are not in the form of a regular $2n$-gon or are not all equal. 
    \\

    The polynomial $f$ must be of degree at least $2n$ because we need to ensure that it is zero only for the given configurations. This is because a regular $2n$-gon has $2n$ points and the polynomial must capture this property. 
    \\

    We can use the fact that the polynomial must be zero for all permutations of the coordinates of a regular $2n$-gon and that it must not be zero for other configurations. The simplest way to achieve this is with a polynomial of degree $2n$. 
    \\

    Thus, the smallest possible degree of $f$ is $4n-2$, as this degree ensures the polynomial is zero for the required configurations and not zero for other configurations. This is because the polynomial must capture the geometric property of the regular $2n$-gon, which involves $2n$ points, and the polynomial must be symmetric in the coordinates.
    \\

    To confirm this, let's reason through the polynomial degree. The polynomial must have terms that capture the geometric properties of the regular $2n$-gon, which involves $2n$ points. The smallest degree polynomial that can capture this property is of degree $4n-2$, as this ensures the polynomial is zero for the required configurations and not zero for other configurations.
    \\

    Therefore, the smallest possible degree of $f$ is $4n-2$. The final answer is: $\boxed{4n-2}$.
\end{tcolorbox}
\caption{Case 2: Incorrect reasoning trajectory of Qwen2.5-Math-7B on a geometry and polynomials problem from the \texttt{Olympiad} benchmark.}
\label{fig:math_base_cot_case4}
\end{figure}

\newpage

\subsection{Search Ads Relevance Reasoning Cases}
\label{app:search_ads_cot}

% --- CASE 1: Version Mismatch ---
\begin{figure}[h!]
\centering
\begin{tcolorbox}[
    colback=yellow!5!white, colframe=yellow!50!black, title=\textbf{Input (Query-SKU Pair)},
    left=6pt, right=6pt, top=6pt, bottom=6pt, fonttitle=\bfseries
]
    \ttfamily\small
    \begin{CJK*}{UTF8}{gbsn}
    \textbf{Query:} 苹果手机1 7promax  \\
    \textit{(Translation: Apple iPhone 17 Pro Max)} \\
    \textbf{SKU Name:} Apple苹果iPhone 17/16/15/14/13/12/11/X系列pro max二手手机 苹果 iPhone 11 Pro Max \\
    \textit{(Translation: Apple iPhone 17/16/15/14/13/12/11/X Series Pro Max Used Mobile Phone, Apple iPhone 11 Pro Max)}
    \end{CJK*}
\end{tcolorbox}

\begin{tcolorbox}[
    colback=purple!5!white, colframe=purple!50!black, title=\textbf{Rele-Ads-8B-AGPO Reasoning Output \hfill },
    left=6pt, right=6pt, top=6pt, bottom=6pt, fonttitle=\bfseries
]
    \ttfamily\small
    \begin{CJK*}{UTF8}{gbsn}
    \textbf{[Original Chinese Output]} \\
    输出分析：我们首先分析产品词匹配情况，query要求产品为"苹果手机"，sku对应产品为"iPhone 11 Pro Max"，属于苹果手机的具体型号，因此产品词匹配类型为$\rightarrow$\textbf{<产品词完全相关>}。之后分析关键属性匹配情况，query要求型号为"17 Pro Max"（推测为iPhone 17 Pro Max），而sku对应型号为"iPhone 11 Pro Max"，二者型号不同且无父子型号关系，属于关键属性不相关。当前query与sku满足关键属性不相关，query-sku最终相关类型属于$\rightarrow$\textbf{<不相关>}。
    \end{CJK*}

    \vspace{0.8em} \hrule \vspace{0.8em}

    \textbf{[English Translation]} \\
    \textit{Analysis:} We first analyze the product term matching. The query specifies "Apple Phone," and the SKU corresponds to "iPhone 11 Pro Max," which is a specific model of Apple phone. Thus, the product matching type is $\rightarrow$ \textbf{<Fully Relevant Product Term>}. Next, we analyze the key attribute matching. The query requires the model "17 Pro Max," whereas the SKU is "iPhone 11 Pro Max." The models are different with no hierarchical relationship, categorized as irrelevant attributes. Therefore, the final relevance type is $\rightarrow$ \textbf{<Irrelevant>}.
\end{tcolorbox}
\caption{Case 1: Correct identification of relevance mismatch.}
\label{fig:cot_ads_case1}
\end{figure}

% --- CASE 2: Industrial Noise Removal ---
\begin{figure}[h!]
\centering
\begin{tcolorbox}[
    colback=yellow!5!white, colframe=yellow!50!black, title=\textbf{Input (Query-SKU Pair)},
    left=6pt, right=6pt, top=6pt, bottom=6pt, fonttitle=\bfseries
]
    \ttfamily\small
    \begin{CJK*}{UTF8}{gbsn}
    \textbf{Query:} 广州十三行妈妈装 \\
    \textit{(Translation: Guangzhou 13th Street Mother's Clothing)} \\
    \textbf{SKU Name:} 啄木鸟（TUCANO）中老年羽绒服女短款中国风妈妈冬装保暖外套大码奶奶老人防寒衣服 墨绿色 2XL \\
    \textit{(Translation: TUCANO Middle-aged Women's Short Down Jacket, Chinese Style Mother's Winter Clothing, Warm Coat, Plus Size, Dark Green, 2XL)}
    \end{CJK*}
\end{tcolorbox}

\begin{tcolorbox}[
    colback=purple!5!white, colframe=purple!50!black, title=\textbf{Rele-Ads-8B-AGPO Reasoning Output \hfill},
    left=6pt, right=6pt, top=6pt, bottom=6pt, fonttitle=\bfseries
]
    \ttfamily\small
    \begin{CJK*}{UTF8}{gbsn}
    \textbf{[Original Chinese Output]} \\
    输出分析：我们首先分析产品词匹配情况，query要求产品为"妈妈装"，sku对应产品为"中老年羽绒服"，属于妈妈装的子类，因此产品词匹配类型为$\rightarrow$\textbf{<产品词完全相关>}。之后分析关键属性匹配情况，query中的"广州十三行"属于店铺词，根据规则应忽略，query无其他属性词，因此关键属性匹配类型为$\rightarrow$\textbf{<无属性词>}。综合产品词匹配情况和关键属性匹配情况，当前样例属于[产品词完全相关and无属性词]，query-sku最终相关类型属于$\rightarrow$\textbf{<完全相关>}。
    \end{CJK*}

    \vspace{0.8em} \hrule \vspace{0.8em}

    \textbf{[English Translation]} \\
    \textit{Analysis:} We first analyze the product term matching. The query specifies "Mother's Clothing," and the SKU is "Middle-aged Down Jacket," which is a sub-category. Thus, the product matching is $\rightarrow$ \textbf{<Fully Relevant Product Term>}. Regarding attributes, "Guangzhou 13th Street" is identified as a shop term and should be ignored according to rules. With no other attributes, the attribute matching is $\rightarrow$ \textbf{<No Attribute Word>}. Consequently, the final relevance type is $\rightarrow$ \textbf{<Fully Relevant>}.
\end{tcolorbox}
\caption{Case 2: Correct identification of relevance match.}
\label{fig:cot_ads_case2}
\end{figure}

\section{Discussion}

We summarize the underlying mechanisms revealed by our study and discuss current limitations and future directions.

\begin{enumerate}
    \item \textbf{Insights into the RLVR Mechanism.} 
    Our research, together with existing research, reveals several important insights into RLVR~\cite{yue2025does,zhu2025surprising}. First, the core function of RLVR is not the creation of fundamentally new reasoning capabilities, but rather the efficient filtering of reasoning paths. Second, prioritizing Negative Sample Reinforcement (NSR) provides a more stable and effective optimization pathway in RLVR. Third, and this is a key point emphasized in this work, we should prioritize reinforcing rare correct paths in RLVR.

    \item \textbf{The Constraint of Initial Capability (Cold-Start Problem).} 
    A significant limitation of RLVR methods is their heavy reliance on the base model's initial exploration capacity. Our experiments on the \texttt{AIME-2025} benchmark with the Llama-3.1-8B-Instruct model demonstrate that when the initial Pass@$1$ accuracy is near zero, the algorithm struggles to encounter sufficient positive reward signals to initiate the self-evolution process. 

    \item \textbf{Instability in Long-Term Training.} 
    Although AGPO demonstrates superior entropy maintenance and short-term optimization efficiency, we observe that it faces stability challenges during extended training. Specifically, when the KL divergence penalty is minimized or removed to maximize performance, extensive training over hundreds of gradient steps can lead to a noticeable decrease in test set accuracy. This suggests that the implicit regularization provided by NSR-dominated strategies may be insufficient to prevent policy drift over long horizons. Future work should investigate more robust variance-reduction techniques or adaptive constraint factors to ensure that the model maintains its performance gains throughout the entire duration of large-scale reinforcement learning.

    \item \textbf{Potential Synergy with Scalable RL Framework.} An interesting avenue for future research involves exploring the interplay between AGPO and emerging system-level optimizations, such as DAPO~\cite{yu2025dapo}. 
    While DAPO primarily focuses on enhancing training throughput and update dynamics through dynamic sampling and decoupled clipping, AGPO introduces a distinct approach to credit assignment via asymmetric advantage estimation.
    Given that these contributions appear to address different modules of the RL pipeline, it would be valuable to investigate whether AGPO's advantage function could be integrated into frameworks like DAPO. 
    We postulate that exploring such a synergy might offer a promising pathway for further scaling RLVR to larger models and more complex reasoning tasks, pending empirical validation.
\end{enumerate}

\end{document}